\documentclass{article}

\usepackage[final,main]{neurips_2026}

\usepackage[utf8]{inputenc} 
\usepackage[T1]{fontenc}    
\usepackage{hyperref}       
\usepackage{url}            
\usepackage{booktabs}       
\usepackage{amsfonts}       
\usepackage{nicefrac}       
\usepackage{microtype}      
\usepackage{xcolor}         

\usepackage{hyperref}
\usepackage{url}
\usepackage{booktabs}
\usepackage{amsfonts}
\usepackage{amsmath}
\usepackage{amssymb}
\usepackage{microtype}
\usepackage{xcolor}
\usepackage{graphicx}
\usepackage{multirow}
\usepackage{array}
\usepackage{pifont}
\usepackage{enumitem}
\usepackage{subcaption}

\usepackage{wrapfig}
\usepackage[table]{xcolor}

\definecolor{rowfullattn}{RGB}{248,248,248}
\definecolor{rowstream}{RGB}{247,250,255}
\definecolor{rowours}{RGB}{247,255,247}
\definecolor{fullattngray}{RGB}{115,115,115}
\newcommand{\gray}[1]{\textcolor{fullattngray}{#1}}

\definecolor{rowfullattn}{RGB}{255,248,238}
\definecolor{rowstream}{RGB}{247,250,255}
\definecolor{rowours}{RGB}{247,255,247}
\definecolor{rowablate}{RGB}{255,255,247}
\definecolor{rowphysical}{RGB}{255,247,247}
\definecolor{fullattngray}{RGB}{115,115,115}

\definecolor{my_darkred}{RGB}{150, 0, 0} 

\newcommand{\inlinepara}[1]{\vspace{0.25em}\noindent\textbf{#1.}}
\definecolor{my_darkblue}{RGB}{0, 0, 150}

\newcommand{\ours}{HDR}

\newcommand{\cmark}{\ding{51}}
\newcommand{\xmark}{\ding{55}}
\newcommand{\meanstd}[2]{\mbox{#1{\scriptsize$\pm$#2}}}

\title{Hierarchical Denoising For Multi-Step Visual Reasoning}

\author{
Zezhong Qian\textsuperscript{1,6},
Xiaowei Chi\textsuperscript{2,6},
Chak-Wing Mak\textsuperscript{1,6},
Tianze Zhou\textsuperscript{3,6},\\
Ruibin Yuan\textsuperscript{2,5},
Yuhan Rui\textsuperscript{1,6},
Hengzhe Sun\textsuperscript{1},
Zhuoqun Wu\textsuperscript{4},\\
Yuming Li\textsuperscript{1},
Siyuan Qian\textsuperscript{1},
Sirui Han\textsuperscript{2},
Shanghang Zhang\textsuperscript{1}
\\[0.5em]
\begin{minipage}{0.95\textwidth}
\centering
\normalfont\small
\textsuperscript{1}State Key Laboratory of Multimedia Information Processing,
School of Computer Science, Peking University\\
\textsuperscript{2}The Hong Kong University of Science and Technology\\
\textsuperscript{3}Beihang University\\
\textsuperscript{4}Fuzhou University\\
\textsuperscript{5}Multimodal Art Projection\\
\textsuperscript{6}Muka Robotics
\end{minipage}
}

\begin{document}

\maketitle

\begin{abstract} 
Video models are recently evolving into vision foundation models, but they still lack human-like, multi-step reasoning.
Existing streaming autoregressive diffusion models are efficient but lack the reasoning ability, whereas bidirectional diffusion allows for global revision but incurs high inference cost due to the dense frames in fixed-sequence denoising. Consequently, both paradigms struggle to maintain logical consistency with low-latency streaming in complex reasoning tasks.
Bridging this gap, we propose \textbf{HDR} (\emph{Hierarchical Denoising for Visual Reasoning}), 
a unified framework for multi-step reasoning by integrating hierarchical latents into the causal video generation process.
HDR organizes video latents into a tree-structured hierarchy to perform coarse-to-fine reasoning before streaming output.
Coarse denoising layers maintain uncertain hypotheses for global planning, while finer denoising layers progressively refine them into concrete visual states. 
A sparse hierarchical attention pattern (SHAP) further reduces temporal attention cost.
We construct a level-stratified multi-step video reasoning benchmark with out-of-distribution cases, covering six tasks:
\textit{maze navigation, Tower of Hanoi, one-line drawing, sliding puzzle, Sokoban, and water pouring}. 
Compared with the streaming autoregressive diffusion baseline, HDR improves overall success from \(34.22\) to \(60.29\) (\(76.2\%\) relative gain) in multi-step reasoning accuracy, and improves average progress from \(76.00\) to \(89.56\), indicating more consistent intermediate reasoning trajectories. For deployment efficiency, HDR maintains low-latency streaming at \(0.70\)s per latent, \(54.2\times\) faster than bidirectional diffusion during streaming. HDR also demonstrates strong data efficiency, retaining \(82.9\%\) of its full-data success score using only \(2\%\) of the training data, compared with \(52.0\%\) for bidirectional diffusion.
Further experiments on real-world robots showcase the potential of HDR in physical interaction, providing a new paradigm for physical world modeling. Project demo page is available at \url{https://hierarchical-diffusion-reasoning.github.io/}.
\end{abstract}

\section{Introduction}

Video models are evolving from realistic video synthesizers into generalist visual foundation models capable of visual reasoning \cite{wiedemer2025videomodelszeroshotlearners,wang2026demystifingvideoreasoning,wang2026bigvideoreasoningsuite,wu2026visualgenerationunlockshumanlike, zhang2025worldmodelsbenefitvlms,yu2024languagemodelbeatsdiffusion, kondratyuk2024videopoetlargelanguagemodel}. Recent work such as Veo3 \cite{wiedemer2025videomodelszeroshotlearners} showed that large video generation models can solve tasks such as maze navigation, symmetry completion, physical reasoning, and tool-use simulation through generated visual trajectories. Complementarily, recent analysis of diffusion-based video reasoning suggests that such reasoning is closely tied to intermediate denoising states, where the model can maintain and refine candidate solutions over multiple steps \cite{wang2026demystifingvideoreasoning}. These findings raise a key question: how can video models support reliable multi-step reasoning while still enabling low-latency streaming generation?

Existing video generation paradigms struggle to satisfy both multi-step reasoning and low-latency streaming. Streaming autoregressive diffusion models, such as CausVid \cite{yin2025slowbidirectionalfastautoregressive} and CausalForcing \cite{zhu2026causalforcingautoregressivediffusion} generate efficiently by conditioning each step only on past context, but this left-to-right commitment limits their ability to revise previous decisions and perform multi-step reasoning~\cite{yan2021videogptvideogenerationusing,deng2025autoregressivevideogenerationvector,huang2025selfforcingbridgingtraintest,liu2025rollingforcingautoregressivelong}. In contrast, bidirectional diffusion models jointly denoise a fixed video sequence, allowing global information flow and revision across time~\cite{nvidia2025cosmosworldfoundationmodel,wan2025wanopenadvancedlargescale,wiedemer2025videomodelszeroshotlearners}. However, this requires dense full-sequence updates at every denoising step, leading to high deployment cost and poor compatibility with streaming generation~\cite{yin2025slowbidirectionalfastautoregressive,gao2025ca2vdmefficientautoregressivevideo,blattmann2023stablevideodiffusionscaling,qian2025wristworldgeneratingwristviews4d}. These limitations reveal a fundamental tension: current models either generate efficiently but struggle with logical consistency over multiple steps, or reason globally at the cost of high-latency fixed-sequence denoising.

\begin{figure*}[t]
    \centering
    \includegraphics[width=\textwidth]{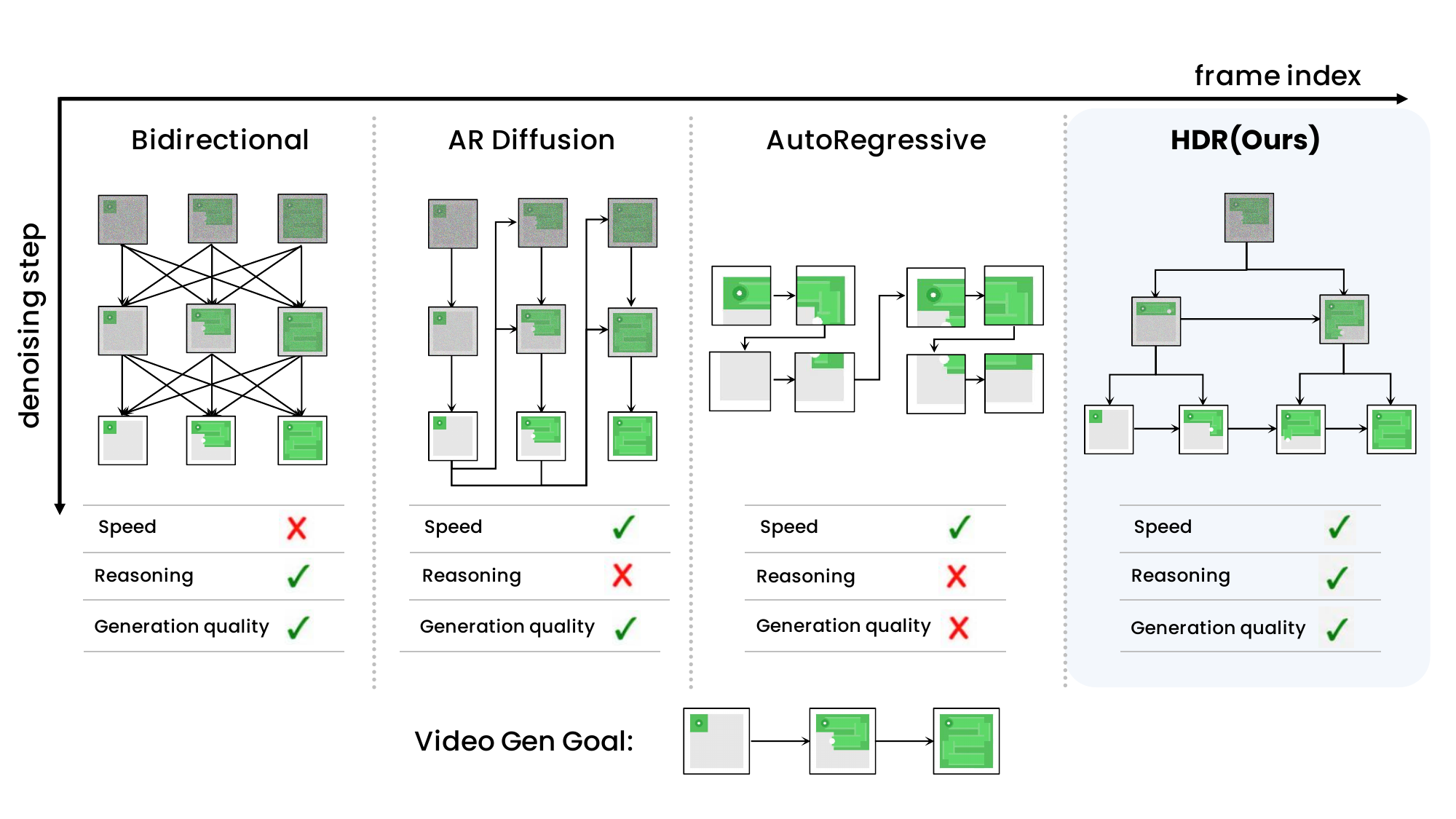}
    \caption{Comparison of video generation paradigms. Streaming autoregressive diffusion models are efficient but commit too early for reliable multi-step reasoning, while bidirectional diffusion supports global revision but requires costly dense fixed-sequence denoising. HDR performs hierarchical denoising before streaming output: coarse layers maintain high-level hypotheses, finer layers refine them into visual states, and sparse hierarchical attention keeps generation efficient.}
     \vspace{-20pt}
    \label{fig:teaser}
\end{figure*}

Motivated by this observation, we propose \textbf{HDR} (\emph{Hierarchical Denoising for Visual Reasoning}), a unified framework that integrates hierarchical latents into the streaming autoregressive diffusion process. As illustrated in Figure~\ref{fig:teaser}, HDR organizes video latents into a tree-structured hierarchy and performs coarse-to-fine multi-step reasoning before streaming output. This hierarchy gives the model an explicit intermediate space for high-level planning before it commits to frame-level generation.

A key design of HDR is to match denoising strength to hierarchy level. Instead of fully denoising every layer, HDR keeps coarse layers at higher noise levels so they can preserve multiple possible global plans, while finer layers receive stronger denoising and lower residual noise to instantiate these plans into concrete visual states. HDR further introduces \textbf{SHAP} (\emph{Sparse Hierarchical Attention Pattern}), which lets each token communicate only with fixed local and parent-level contexts for fast retrieval. This enables multi-scale information flow without dense full-sequence attention, reducing temporal attention cost while preserving streaming generation.


We construct a level-stratified multi-step video reasoning benchmark with out-of-distribution cases, covering six tasks: \textit{maze navigation, Tower of Hanoi, one-line drawing, sliding puzzle, Sokoban, and water pouring}. These tasks require models to maintain logical consistency across multi-step reasoning trajectories rather than merely generating locally plausible motion. Experiments show that HDR substantially improves both final task completion and intermediate reasoning consistency: it improves the overall success score from \(34.22\) to \(60.29\) (\(76.2\%\) relative gain), and increases the overall average progress score from \(76.00\) to \(89.56\). HDR also preserves efficient streaming behavior, achieving \(0.70\)s per latent during streaming, \(54.2\times\) faster than bidirectional diffusion. In addition, HDR demonstrates strong data efficiency, retaining \(82.9\%\) of its full-data success score when trained with only \(2\%\) of the data, compared with \(52.0\%\) for bidirectional diffusion. Finally, real-world robot maze experiments show that HDR can transfer its hierarchical reasoning ability to physical interaction, suggesting its potential as a new paradigm for physical world modeling.

Our contributions can be summarized as follows:
\begin{itemize}[leftmargin=1.2em,itemsep=0.15em,topsep=0.2em,parsep=0em,partopsep=0em]
    \item We identify the core tension in multi-step video reasoning: models must maintain logical consistency across long trajectories while also supporting low-latency streaming generation.
    \item We propose \textbf{HDR} (\emph{Hierarchical Denoising for Visual Reasoning}), a unified framework that integrates hierarchical latents into streaming autoregressive diffusion and performs coarse-to-fine reasoning before streaming output.
   \item We introduce a hierarchy-matched denoising schedule and \textbf{SHAP} (\emph{Sparse Hierarchical Attention Pattern}). The former preserves high-level hypotheses at coarse layers and refines them at finer layers, while the latter reduces temporal attention cost through local and parent-level contexts.
    \item We construct a level-stratified multi-step video reasoning benchmark with OOD cases, and demonstrate HDR's advantages in reasoning accuracy, data efficiency, low-latency streaming, and physical-world robot interaction.
\end{itemize}

\section{Related Work}

\inlinepara{Video Model Reasoning}
Recent studies show that video generation models can exhibit reasoning behaviors beyond visual realism. Benchmarks such as VBVR, V-ReasonBench, and VR-Bench evaluate these capabilities across structured problem solving, spatial cognition, physical dynamics, maze navigation, and multi-step planning~\cite{wang2026bigvideoreasoningsuite,luo2025vreasonbenchunifiedreasoningbenchmark,yu2025vrbenchbenchmarkmultistepreasoning}. Unlike video understanding, where the input video is fixed, video generation requires the model to construct a coherent future trajectory that satisfies both local visual plausibility and global task constraints. This makes generated videos a natural interface for world-model-style reasoning, but also makes early visualized mistakes difficult to correct~\cite{wu2026visualgenerationunlockshumanlike}. Recent analyses further suggest that reasoning in video diffusion models is closely tied to iterative denoising dynamics, where intermediate latents maintain and refine uncertain hypotheses, rather than arising solely from frame-by-frame prediction~\cite{wang2026demystifingvideoreasoning}. However, existing work mainly benchmarks or analyzes emergent reasoning in bidirectional generators, rather than designing architectures that preserve reasoning ability under streaming generation constraints~\cite{wang2026bigvideoreasoningsuite,luo2025vreasonbenchunifiedreasoningbenchmark,wang2026demystifingvideoreasoning}.

\inlinepara{Autoregressive Video Diffusion Models}
Streaming autoregressive diffusion models replace dense full-sequence denoising with sequential temporal computation, enabling low-latency video generation and efficient KV-cache reuse. Recent work improves this paradigm through chunk-wise rollout, queue-based denoising, AR-guided diffusion, causal attention, training--inference alignment, distillation, cache sharing, and sliding-window KV-cache acceleration~\cite{henschel2025streamingt2vconsistentdynamicextendable,kim2024fifodiffusiongeneratinginfinitevideos,li2025arlonboostingdiffusiontransformers,yin2025slowbidirectionalfastautoregressive,gao2025ca2vdmefficientautoregressivevideo,huang2025selfforcingbridgingtraintest,zhu2026causalforcingautoregressivediffusion,yang2025longliverealtimeinteractivelong,robbyantteam2026advancingopensourceworldmodels,pyramid,jia2025ditar,xiao2024efficient}. These properties make autoregressive video models attractive for closed-loop robot interaction \cite{li2026causalworldmodelingrobot,ye2026worldactionmodelszeroshot}, where low latency is critical, and their context-as-memory structure allows previously generated visual states to serve as a persistent temporal memory \cite{yu2025contextmemorysceneconsistentinteractive,hong2025relicinteractivevideoworld}. Moreover, because generation proceeds autoregressively, sliding-window KV-cache mechanisms can extend rollout length and support effectively unbounded video generation \cite{yang2025longliverealtimeinteractivelong,liu2025rollingforcingautoregressivelong,gao2025ca2vdmefficientautoregressivevideo}. However, the same sequential commitment makes earlier decisions difficult to revise: once a frame or latent chunk has been produced, later predictions condition on this committed history. HDR preserves the low-latency structure of streaming autoregressive diffusion while introducing a tree-structured latent hierarchy for coarse-to-fine denoising, enabling revisable high-level planning before committing to fine-grained visual details.

\section{Method}

\begin{figure*}[t]
    \centering
    \includegraphics[width=\textwidth]{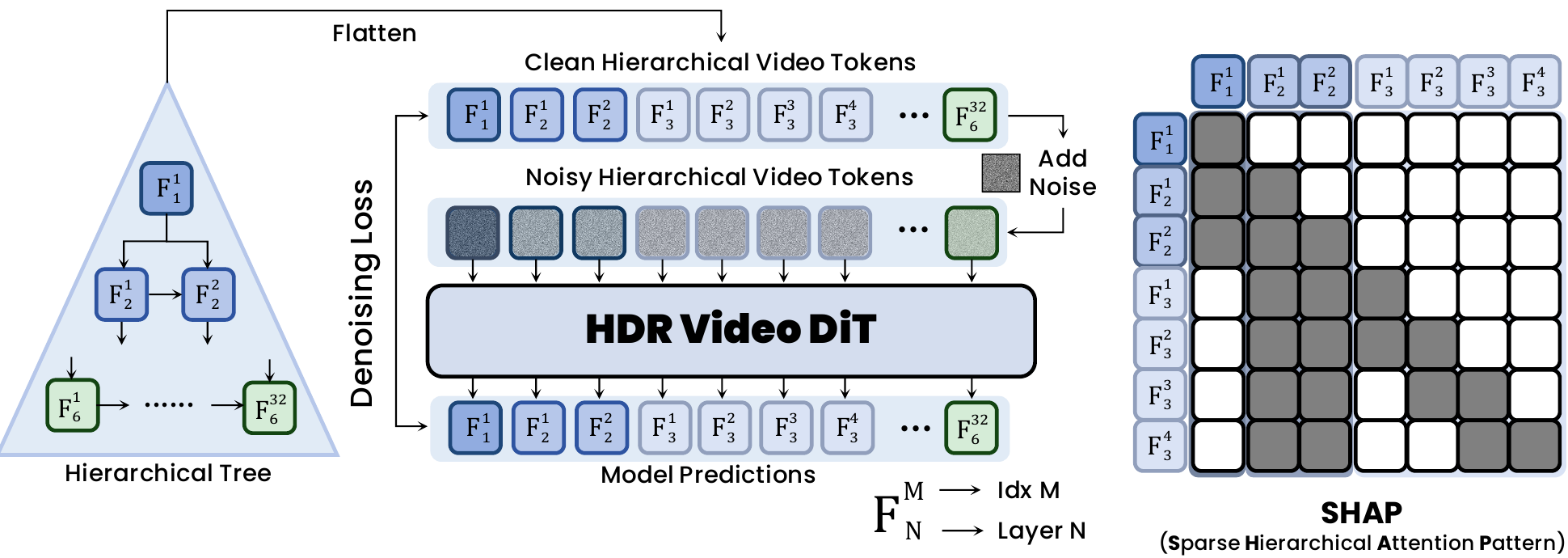}
    \caption{Overview of HDR. Video latents are organized into a tree-structured hierarchy across multiple temporal resolutions. During training, each hierarchy token is corrupted along a flow-matching path and HDR Video DiT is optimized with a layer-wise objective. During inference, all tree tokens are flattened into a coarse-to-fine autoregressive order. \textbf{SHAP} (\emph{Sparse Hierarchical Attention Pattern}) defines a structured attention mask over this flattened sequence: each token attends only to fixed local, parent-level, and first-frame contexts rather than the full video sequence. Generated tokens are written into a shared KV cache and reused by later tokens across hierarchy levels, enabling multi-scale information propagation with low temporal attention cost.}
    \vspace{-15pt}
    \label{fig:main}
\end{figure*}

We present \ours{} (\emph{Hierarchical Denoising for Visual Reasoning}), a hierarchical framework for multi-step video reasoning. HDR preserves the low-latency streaming behavior of streaming autoregressive diffusion while introducing a structured intermediate process for global planning and revision. We first revisit why streaming autoregressive generation struggles with multi-step reasoning, then introduce the hierarchical latent representation, layer-wise flow-matching objective, and \textbf{SHAP} (\emph{Sparse Hierarchical Attention Pattern}), which enables efficient inference over flattened tree tokens.

\subsection{Rethinking Streaming Autoregressive Generation for Multi-step Reasoning}

We start by comparing bidirectional diffusion and streaming autoregressive diffusion. Let \(\mathbf{z}^{t}=\{z_1^{t},\ldots,z_N^{t}\}\) denote a video latent sequence at denoising step \(t\), where \(N\) is the number of temporal latent tokens and \(z_i^{t}\) is the token at temporal position \(i\). Let \(c\) denote the conditioning signal, such as text, image, or the first frame. A bidirectional video diffusion model updates the entire sequence jointly at every denoising step~\cite{wan2025wanopenadvancedlargescale,nvidia2025cosmosworldfoundationmodel}:
\begin{equation}
\mathbf{z}^{t-1}
=
D_{\theta}(\mathbf{z}^{t}, t, c),
\label{eq:bidirectional_update}
\end{equation}
where \(D_{\theta}\) is the denoising network. Because all temporal tokens remain noisy during intermediate denoising steps, information can propagate across the whole sequence before video is committed. This allows uncertain hypotheses to be maintained and refined over multiple denoising steps~\cite{wang2026demystifingvideoreasoning}. However, the same global revision ability comes with high cost: each step repeatedly updates a dense fixed-length sequence, making bidirectional diffusion poorly aligned with low-latency streaming.

AR diffusion improves deployment efficiency by factorizing generation from left to right:
\begin{equation}
p(\mathbf{z}\mid c)
=
\prod_{i=1}^{N}
p(z_i \mid z_{<i}, c),
\label{eq:streaming_ar_factorization}
\end{equation}
where \(\mathbf{z}=\{z_1,\ldots,z_N\}\) is the clean latent sequence and \(z_{<i}\) denotes all previously generated temporal tokens. With an autoregressive attention mask, token \(z_i\) attends only to past tokens and the condition \(c\), enabling streaming inference and KV-cache reuse~\cite{yin2025slowbidirectionalfastautoregressive,zhu2026causalforcingautoregressivediffusion,huang2025selfforcingbridgingtraintest}. However, this structure also creates an irreversible rollout: once \(z_i\) is generated, future predictions follow \(z_i \rightarrow p(z_{i+1}\mid z_{\leq i},c) \rightarrow p(z_{i+2}\mid z_{\leq i+1},c) \rightarrow \cdots\). If an early token encodes an incorrect decision, later tokens must condition on this committed history and cannot revise it, which weakens logical consistency across multi-step reasoning trajectories.

Thus, the central challenge is not simply choosing between bidirectional and streaming autoregressive generation. Bidirectional diffusion supports global revision but incurs high deployment cost, while streaming autoregressive diffusion is efficient but lacks a mechanism for revisable multi-step reasoning. HDR addresses this tension by introducing a hierarchy of latent variables: coarse levels preserve noisy high-level hypotheses for global planning, while finer levels progressively refine them into concrete visual states before streaming output.

\subsection{Hierarchical Denoising for Visual Reasoning}

HDR represents a video using a tree-structured hierarchy of latent tokens:
\begin{equation}
\mathcal{T}
=
\{\mathcal{V}^{1},\mathcal{V}^{2},\ldots,\mathcal{V}^{L}\},
\qquad
\mathcal{V}^{\ell}
=
\{v_{\ell,1},\ldots,v_{\ell,N_\ell}\}.
\label{eq:hierarchy_definition}
\end{equation}
Here, \(L\) is the number of hierarchy levels, \(\mathcal{V}^{\ell}\) is the set of latent tokens at level \(\ell\), \(N_\ell\) is the number of tokens at that level, and \(v_{\ell,i}\) denotes the \(i\)-th token. We use \(\ell=1\) for the coarsest level and \(\ell=L\) for the finest level. Coarse tokens summarize global temporal structure and represent high-level plans, while fine tokens encode local visual details and instantiate the final video dynamics. Each non-root token has a parent in the previous coarser level, denoted as \(\pi(\ell,i)=(\ell-1,p_\ell(i))\) for \(\ell>1\), where \(p_\ell(i)\) maps token \(v_{\ell,i}\) to its parent index at level \(\ell-1\). The parent token represents the coarse temporal segment that the current token refines.

As shown in Figure~\ref{fig:main}, HDR constructs hierarchical tokens from the input video latent sequence and performs coarse-to-fine reasoning before streaming output. During training, each hierarchy token is corrupted along a flow-matching path, and the model learns to predict the corresponding velocity target under hierarchical context. During inference, HDR generates tokens from coarse to fine: upper layers form noisy but revisable global hypotheses, while lower layers refine these hypotheses into concrete visual states. Within each level, generation proceeds autoregressively from left to right, preserving the streaming structure of autoregressive diffusion.

\subsection{Layer-wise Flow-Matching Objective}

We now describe the training objective over hierarchical tokens. For a clean hierarchy token \(v_{\ell,i}^{0}\), we sample a noise token \(\epsilon\sim\mathcal{N}(0,I)\) and construct an interpolated token at continuous time \(t\in[0,1]\) as \(v_{\ell,i}^{t}=(1-t)v_{\ell,i}^{0}+t\epsilon\). Under this linear probability path, the target velocity field is \(u_{\ell,i}^{t}=\epsilon-v_{\ell,i}^{0}\). The HDR network is trained to predict this flow velocity from the interpolated token, the timestep, the hierarchical context \(h_{\ell,i}\), and the condition \(c\). The layer-wise flow-matching objective sums the velocity regression loss over all hierarchy levels and tokens:
\begin{equation}
\mathcal{L}_{\mathrm{HDR}}
=
\sum_{\ell=1}^{L}
\lambda_\ell
\frac{1}{N_\ell}
\sum_{i=1}^{N_\ell}
\mathbb{E}_{t,\epsilon}
\left[
\left\|
\left(\epsilon - v_{\ell,i}^{0}\right)
-
u_{\theta}
\left(
v_{\ell,i}^{t},
t,
h_{\ell,i},
c
\right)
\right\|_2^2
\right].
\label{eq:hdr_loss}
\end{equation}
Here, \(h_{\ell,i}\) denotes the sparse hierarchical context available to token \(v_{\ell,i}\), and \(\lambda_\ell\) balances the contribution of different hierarchy levels. This objective encourages each level to learn a velocity field appropriate to its temporal abstraction: coarse levels model global planning structure, while fine levels model concrete visual states and motion details.

A key design of HDR is to match denoising strength to hierarchy level. Instead of assigning the same inference budget to every layer, HDR uses a level-dependent sampling budget \(K_\ell\), with \(K_1 < K_2 < \cdots < K_L\). Equivalently, the residual noise level decreases from coarse to fine, i.e., \(\rho_1 > \rho_2 > \cdots > \rho_L\). Coarse layers are intentionally stopped at higher noise levels, so their predictions remain partially stochastic and can preserve multiple possible global plans. Finer layers receive stronger denoising and lower residual noise, progressively instantiating these hypotheses into visual states. We provide the entropy-matched derivation of \(K_\ell\) and its ablation in Appendix~\ref{sec:entropy_matched}.

\subsection{Sparse Hierarchical Attention Pattern}

HDR implements hierarchical reasoning with \textbf{SHAP} (\emph{Sparse Hierarchical Attention Pattern}), a structured attention mask over flattened tree tokens. Each hierarchy token is indexed by \((\ell,i)\), where \(\ell\) is the hierarchy level and \(i\) is the temporal position within that level. To perform inference autoregressively, we flatten all tree tokens into a coarse-to-fine order using \(\phi(\ell,i)=i+\sum_{r<\ell}N_r\), with \(s=\phi(\ell,i)\) denoting the flattened token index. Thus, all tokens in \(\mathcal{V}^{1}\) are generated first, followed by \(\mathcal{V}^{2}\), and so on until the finest level \(\mathcal{V}^{L}\). This ordering matches the inference path shown in Figure~\ref{fig:main}: higher-level tokens are generated before the lower-level tokens that refine them.

For token \(v_{\ell,i}\), SHAP first defines its sparse context in tree coordinates:
\begin{equation}
\begin{aligned}
\mathcal{A}_{\mathrm{SHAP}}(\ell,i)
={}&
\mathbf{1}[i>1]\{(\ell,i-1)\}
\cup
\mathbf{1}[i=1]\{x_{\mathrm{ref}}\} \\
&\cup
\mathbf{1}[\ell>1]
\{\pi(\ell,i),\mathrm{left}(\pi(\ell,i)),\mathrm{right}(\pi(\ell,i))\}.
\end{aligned}
\label{eq:shap_tree_context}
\end{equation}
Here, \(x_{\mathrm{ref}}\) is the clean first-frame condition, \((\ell,i-1)\) provides same-level autoregressive continuity, and \(\pi(\ell,i)\) is the parent token in the coarser level. The neighboring parent-level tokens \(\mathrm{left}(\pi(\ell,i))\) and \(\mathrm{right}(\pi(\ell,i))\) provide boundary information from adjacent coarse segments. Invalid boundary indices are omitted. For root-level tokens, which have no parent, the hierarchical context reduces to the first-frame condition and same-level autoregressive context.

The tree context is then converted into a binary attention mask over the flattened sequence. Let \(S=\sum_{\ell=1}^{L}N_\ell\) be the total number of hierarchy tokens and \(M_{\mathrm{SHAP}}\in\{0,1\}^{S\times S}\) be the SHAP mask. For \(s=\phi(\ell,i)\) and \(r=\phi(\ell',j)\), we define:
\begin{equation}
M_{\mathrm{SHAP}}[s,r]
=
\mathbf{1}
\left[
(\ell',j)\in\mathcal{A}_{\mathrm{SHAP}}(\ell,i)
\right].
\label{eq:shap_mask}
\end{equation}
This mask is sparse by construction: each row contains only a constant number of valid attention targets, independent of video length. It is also autoregressive under the flattened order, because every valid context token is either a previous same-level token, a coarser-level token that has already been generated, or the first-frame condition.

SHAP naturally induces cross-level KV-cache sharing. During inference, once token \(v_{\ell,i}\) is generated, its key-value state is written into a global hierarchy cache, denoted by \(\mathcal{C}_{s}=\mathcal{C}_{s-1}\cup\{(k_{\ell,i},v_{\ell,i}^{\mathrm{KV}})\}\), where \(s=\phi(\ell,i)\). When generating a later token \(v_{\ell',j}\), HDR retrieves only the cached states selected by the SHAP mask, i.e., \(h_{\ell',j}=\mathrm{Attn}(q_{\ell',j},\{(k_{\ell,i},v_{\ell,i}^{\mathrm{KV}}): M_{\mathrm{SHAP}}[\phi(\ell',j),\phi(\ell,i)]=1\})\). Thus, information generated at coarse levels is reused by lower levels through the shared KV cache, while local temporal continuity is preserved by same-level autoregressive links. Because each token attends to a fixed-size SHAP context rather than dense full-sequence tokens, HDR reduces temporal attention cost while maintaining multi-scale information flow before streaming output.

\section{Experiments}
\label{sec:experiment}

We evaluate HDR along two axes: multi-step reasoning ability and low-latency streaming generation efficiency. Section~\ref{sec:benchmarks_metrics} introduces the benchmark, metrics,  baselines, and implementation setup. Section~\ref{sec:main_results} compares HDR with full-attention and streaming baselines. Section~\ref{sec:layer_analysis} analyzes the hierarchy layers impact, Section~\ref{sec:robustness} tests robustness under reduced denoising budgets and limited data, and Section~\ref{sec:physical_world} evaluates transfer to real-world robot interaction.

\subsection{Benchmarks, Baselines, and Implementation Details}
\label{sec:benchmarks_metrics}

\inlinepara{Benchmarks and metrics}
Existing video reasoning benchmarks are not fully suitable for evaluating streaming multi-step generation: many do not release complete evaluation scripts, and most emphasize short clips or perceptual consistency rather than logical consistency across multiple reasoning steps. We therefore construct a controlled benchmark suite with six tasks: Tower of Hanoi, maze navigation, one-line drawing, sliding puzzle, Sokoban, and water pouring. The benchmark is stratified by difficulty and includes OOD cases to test rule transfer beyond frequent training patterns. The scored evaluation set contains \(370\) held-out videos. Each task is evaluated with \emph{success}, which measures exact task completion, and \emph{average progress}, which measures partial progress and reflects intermediate logical consistency. We report task results as \textit{Success / Avg.\ Progress} pairs and compute overall performance as an unweighted average across the six tasks. Full benchmark construction and task-specific evaluation details are provided in Appendix~\ref{sec:benchmark_eval_details}.

\begin{figure*}[t]
    \centering
    \includegraphics[width=\textwidth]{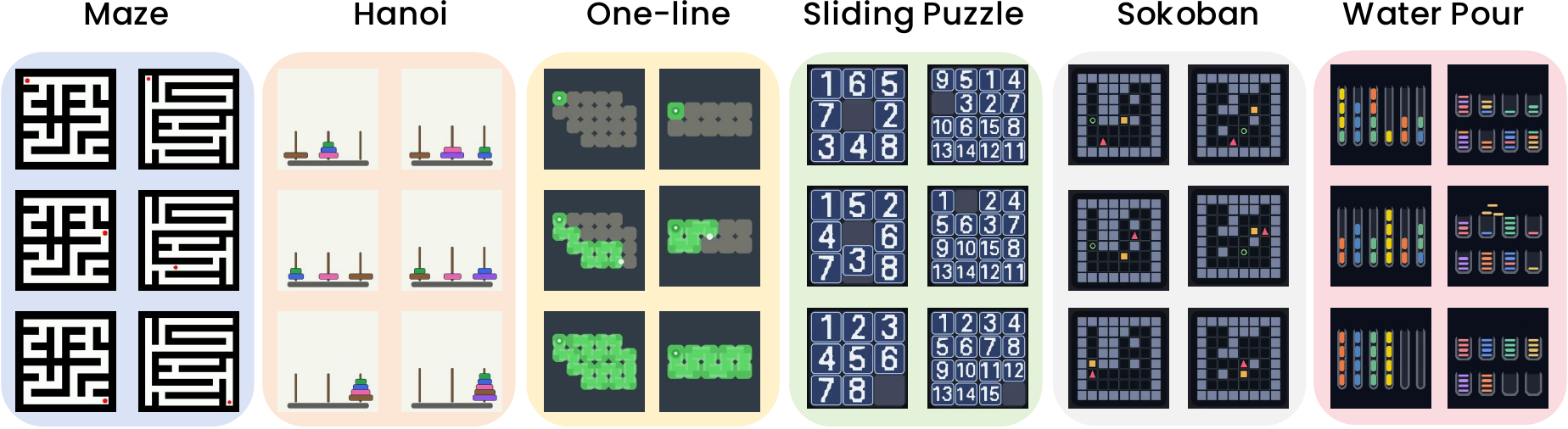}
    \caption{Visualization of the six multi-step video reasoning benchmarks: maze navigation, Tower of Hanoi, one-line drawing, sliding puzzle, Sokoban, and water pouring. Each task requires logical consistency across multiple reasoning steps rather than only local visual plausibility.}
    \label{fig:dataset_vis}
    \vspace{-10pt}
\end{figure*}

\inlinepara{Baselines and implementation}
We compare HDR with full-attention baselines, including bidirectional diffusion and VideoMAE~\cite{tong2022videomaemaskedautoencodersdataefficient}, and streaming baselines, including CausalForcing~\cite{zhu2026causalforcingautoregressivediffusion} and VideoGPT~\cite{yan2021videogptvideogenerationusing}. For a controlled comparison, the main diffusion baselines and HDR are built on Wan2.2-5B-TI2V~\cite{wan2025wanopenadvancedlargescale}; HDR uses six latent hierarchy levels with the entropy-matched denoising schedule \([5,8,13,20,32,50]\). All methods are trained on the same \(18{,}000\)-video reasoning dataset, conditioned on the first frame, and optimized with the same flow-matching training setup. Complete baseline definitions, training details, and implementation settings are provided in Appendix~\ref{sec:baselines_implementation_details}.

\begin{table*}[t]
\centering
\small
\caption{Main comparison on multi-step video reasoning benchmarks. Scores are reported as mean $\pm$ standard deviation for success and average progress. The best and second-best means are shown in \textbf{bold} and \underline{underlined}, respectively. Full-attention baselines are grayed out. Compared with CausalForcing, \ours{} improves overall success from 34.22 to 60.29 and average progress from 76.00 to 89.56.}
\setlength{\tabcolsep}{3.5pt}
\renewcommand{\arraystretch}{1.15}
\resizebox{\linewidth}{!}{%
\begin{tabular}{l c c c c c c c c c}
\toprule
Method & Full Attn. & Metric & Hanoi & Maze & One-line & Sliding & Sokoban & Water & Overall \\
\midrule\midrule

\multirow[c]{2}{*}{\gray{VideoMAE \cite{tong2022videomaemaskedautoencodersdataefficient}}}
& \multirow[c]{2}{*}{\gray{\cmark}}
& \cellcolor{rowfullattn}\gray{Success}
& \cellcolor{rowfullattn}\gray{\meanstd{22.50}{6.18}}
& \cellcolor{rowfullattn}\gray{\meanstd{52.00}{8.96}}
& \cellcolor{rowfullattn}\gray{\meanstd{58.33}{9.67}}
& \cellcolor{rowfullattn}\gray{\meanstd{1.67}{0.00}}
& \cellcolor{rowfullattn}\gray{\meanstd{63.33}{9.42}}
& \cellcolor{rowfullattn}\gray{\textbf{\meanstd{43.33}{4.94}}}
& \cellcolor{rowfullattn}\gray{\meanstd{40.53}{6.53}} \\

&
& \cellcolor{rowfullattn}\gray{Avg.\ Progress}
& \cellcolor{rowfullattn}\gray{\meanstd{58.23}{3.80}}
& \cellcolor{rowfullattn}\gray{\meanstd{93.40}{1.32}}
& \cellcolor{rowfullattn}\gray{\meanstd{91.92}{1.71}}
& \cellcolor{rowfullattn}\gray{\meanstd{49.93}{0.00}}
& \cellcolor{rowfullattn}\gray{\textbf{\meanstd{100.00}{0.00}}}
& \cellcolor{rowfullattn}\gray{\textbf{\meanstd{73.26}{2.48}}}
& \cellcolor{rowfullattn}\gray{\meanstd{77.79}{1.55}} \\
\midrule

\multirow[c]{2}{*}{\gray{Bidirectional \cite{wan2025wanopenadvancedlargescale}}}
& \multirow[c]{2}{*}{\gray{\cmark}}
& \cellcolor{rowfullattn}\gray{Success}
& \cellcolor{rowfullattn}\gray{\underline{\meanstd{45.00}{6.78}}}
& \cellcolor{rowfullattn}\gray{\textbf{\meanstd{90.00}{3.12}}}
& \cellcolor{rowfullattn}\gray{\textbf{\meanstd{81.67}{7.60}}}
& \cellcolor{rowfullattn}\gray{\underline{\meanstd{31.67}{6.98}}}
& \cellcolor{rowfullattn}\gray{\textbf{\meanstd{90.00}{4.21}}}
& \cellcolor{rowfullattn}\gray{\meanstd{21.67}{6.83}}
& \cellcolor{rowfullattn}\gray{\underline{\meanstd{60.00}{5.92}}} \\

&
& \cellcolor{rowfullattn}\gray{Avg.\ Progress}
& \cellcolor{rowfullattn}\gray{\underline{\meanstd{73.58}{2.95}}}
& \cellcolor{rowfullattn}\gray{\textbf{\meanstd{99.89}{0.11}}}
& \cellcolor{rowfullattn}\gray{\textbf{\meanstd{99.26}{0.27}}}
& \cellcolor{rowfullattn}\gray{\underline{\meanstd{93.00}{1.56}}}
& \cellcolor{rowfullattn}\gray{\textbf{\meanstd{100.00}{0.00}}}
& \cellcolor{rowfullattn}\gray{\meanstd{57.08}{3.66}}
& \cellcolor{rowfullattn}\gray{\underline{\meanstd{87.13}{1.42}}} \\

\midrule\midrule

\multirow[c]{2}{*}{VideoGPT \cite{yan2021videogptvideogenerationusing}}
& \multirow[c]{2}{*}{\xmark}
& \cellcolor{rowstream}Success
& \cellcolor{rowstream}\meanstd{18.75}{5.70}
& \cellcolor{rowstream}\meanstd{22.00}{12.83}
& \cellcolor{rowstream}\meanstd{31.67}{7.73}
& \cellcolor{rowstream}\meanstd{10.00}{5.57}
& \cellcolor{rowstream}\meanstd{15.00}{5.46}
& \cellcolor{rowstream}\meanstd{8.33}{3.00}
& \cellcolor{rowstream}\meanstd{17.57}{6.71} \\

&
& \cellcolor{rowstream}Avg.\ Progress
& \cellcolor{rowstream}\meanstd{25.82}{5.46}
& \cellcolor{rowstream}\meanstd{26.32}{12.30}
& \cellcolor{rowstream}\meanstd{81.74}{2.34}
& \cellcolor{rowstream}\meanstd{27.81}{4.01}
& \cellcolor{rowstream}\meanstd{23.50}{4.91}
& \cellcolor{rowstream}\meanstd{38.03}{3.56}
& \cellcolor{rowstream}\meanstd{36.88}{5.43} \\
\midrule

\multirow[c]{2}{*}{CausalForcing \cite{zhu2026causalforcingautoregressivediffusion}}
& \multirow[c]{2}{*}{\xmark}
& \cellcolor{rowstream}Success
& \cellcolor{rowstream}\underline{\meanstd{45.00}{6.73}}
& \cellcolor{rowstream}\meanstd{12.00}{8.08}
& \cellcolor{rowstream}\meanstd{48.33}{7.66}
& \cellcolor{rowstream}\meanstd{21.67}{5.16}
& \cellcolor{rowstream}\meanstd{40.00}{10.69}
& \cellcolor{rowstream}\underline{\meanstd{38.33}{5.25}}
& \cellcolor{rowstream}\meanstd{34.22}{7.26} \\

&
& \cellcolor{rowstream}Avg.\ Progress
& \cellcolor{rowstream}\meanstd{70.47}{2.76}
& \cellcolor{rowstream}\meanstd{55.04}{8.36}
& \cellcolor{rowstream}\meanstd{95.15}{1.28}
& \cellcolor{rowstream}\meanstd{86.13}{2.62}
& \cellcolor{rowstream}\meanstd{82.25}{5.79}
& \cellcolor{rowstream}\meanstd{66.94}{3.03}
& \cellcolor{rowstream}\meanstd{76.00}{3.97} \\
\midrule

\multirow[c]{2}{*}{\textbf{\ours{}}}
& \multirow[c]{2}{*}{\xmark}
& \cellcolor{rowours}Success
& \cellcolor{rowours}\textbf{\meanstd{58.75}{4.50}}
& \cellcolor{rowours}\underline{\meanstd{78.00}{5.25}}
& \cellcolor{rowours}\underline{\meanstd{70.00}{10.12}}
& \cellcolor{rowours}\textbf{\meanstd{33.33}{7.93}}
& \cellcolor{rowours}\underline{\meanstd{78.33}{9.67}}
& \cellcolor{rowours}\textbf{\meanstd{43.33}{4.79}}
& \cellcolor{rowours}\textbf{\meanstd{60.29}{7.04}} \\

&
& \cellcolor{rowours}Avg.\ Progress
& \cellcolor{rowours}\textbf{\meanstd{79.62}{3.04}}
& \cellcolor{rowours}\underline{\meanstd{97.18}{1.07}}
& \cellcolor{rowours}\underline{\meanstd{97.84}{0.61}}
& \cellcolor{rowours}\textbf{\meanstd{93.69}{1.47}}
& \cellcolor{rowours}\underline{\meanstd{99.69}{0.31}}
& \cellcolor{rowours}\underline{\meanstd{69.34}{2.84}}
& \cellcolor{rowours}\textbf{\meanstd{89.56}{1.56}} \\

\bottomrule
\end{tabular}}
\vspace{-10pt}
\label{tab:main}
\end{table*}

\subsection{Main Results: Bridging Reasoning Precision and Streaming Efficiency}
\label{sec:main_results}

\begin{figure*}[t]
    \centering
    \includegraphics[width=\textwidth]{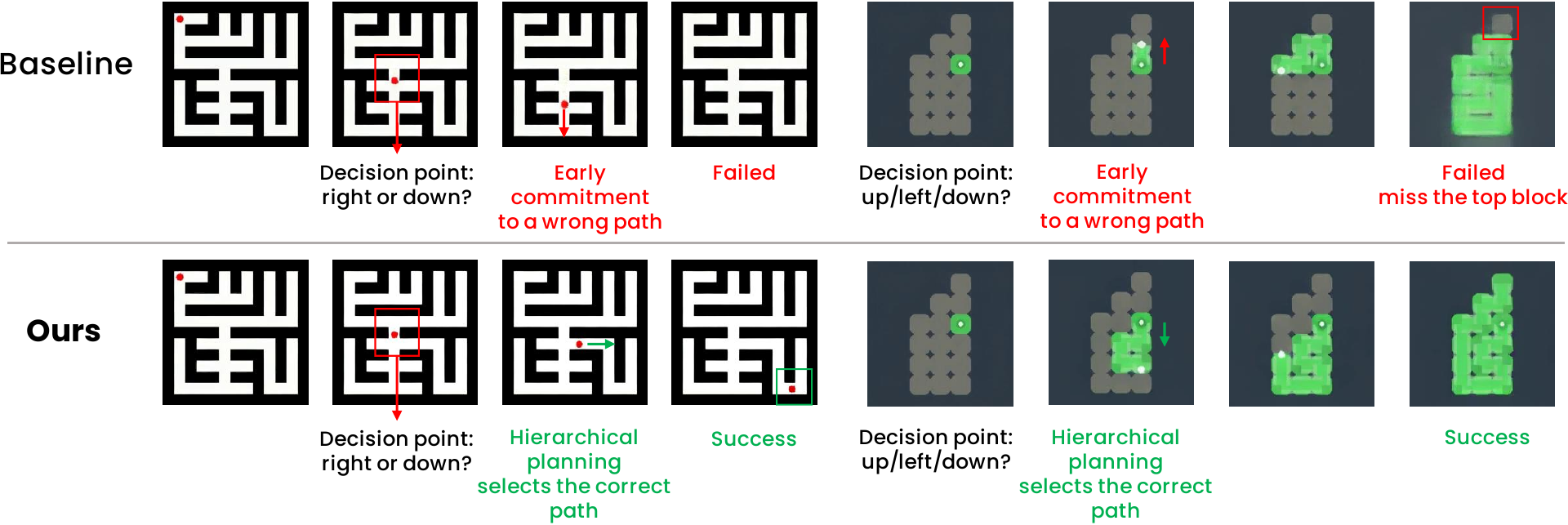}
    \caption{Qualitative comparison between Baseline (CausalForcing) and HDR on Maze and One-line tasks. CausalForcing makes an early local commitment that leads to failure, while HDR performs hierarchical planning before committing to the final trajectory.}
    \label{fig:vis_comparison}
    \vspace{-10pt}
\end{figure*}

HDR improves both the final success of multi-step reasoning trajectories and their intermediate logical consistency. We support this conclusion through quantitative comparisons in Table~\ref{tab:main} and qualitative examples in Figure~\ref{fig:vis_comparison}. We further evaluate streaming efficiency in Table~\ref{tab:inference_speed}, showing that HDR maintains comparable  latency to AR diffusion baseline \cite{zhu2026causalforcingautoregressivediffusion}.

\inlinepara{Overall comparison}
Table~\ref{tab:main} presents the main comparison on our multi-step video reasoning benchmark. Compared with CausalForcing, the streaming autoregressive diffusion baseline, HDR improves overall success from \(34.22\) to \(60.29\), corresponding to a \(76.2\%\) relative gain. Average progress also increases from \(76.00\) to \(89.56\), indicating stronger intermediate logical consistency. Full-attention baselines, shown in gray, enable dense global interaction but are less aligned with low-latency streaming. Despite not using full temporal attention, HDR achieves the best overall success and average progress, showing that hierarchical denoising can recover strong reasoning precision while preserving the streaming structure of autoregressive diffusion.

\begin{wraptable}{r}{0.36\textwidth}
\vspace{-1.2em}
\centering
\caption{Inference speed comparison. Latency is the average time required for each streaming generation step after KV-cache initialization.}
\label{tab:inference_speed}
\small
\setlength{\tabcolsep}{10pt}
\renewcommand{\arraystretch}{1.1}
\resizebox{\linewidth}{!}{%
\begin{tabular}{l c}
\toprule
Method & Latency \\
\midrule
\rowcolor{rowfullattn}
Bidirectional & 37.92s \\
\rowcolor{rowstream}
CausalForcing~\cite{zhu2026causalforcingautoregressivediffusion} & 0.72s \\
\rowcolor{rowours}
\ours{} & \textbf{0.70s} \\
\bottomrule
\end{tabular}}
\vspace{-1.5em}
\end{wraptable}

\inlinepara{Qualitative evidence of revisable planning}
Figure~\ref{fig:vis_comparison} compares CausalForcing and HDR on Maze and One-line cases. CausalForcing commits to an early local decision that leads to failure: taking the wrong branch in Maze or missing the top block in One-line. HDR instead resolves the ambiguous decision point through hierarchical planning before committing to fine-grained outputs, leading to successful task completion.

\inlinepara{Streaming efficiency}
We further report inference speed in Table~\ref{tab:inference_speed}. Latency measures the time required for each streaming generation step after KV-cache initialization. HDR achieves comparable latency to CausalForcing, \(0.70\)s versus \(0.72\)s, while being substantially faster than bidirectional diffusion at \(37.92\)s. This shows that HDR improves multi-step reasoning while preserving low-latency streaming behavior.

\subsection{Mechanism Analysis: Layer-wise Analysis}
\label{sec:layer_analysis}

\begin{figure*}[t]
\centering
\begin{minipage}[c]{0.58\textwidth}
    \centering
    \includegraphics[width=\linewidth]{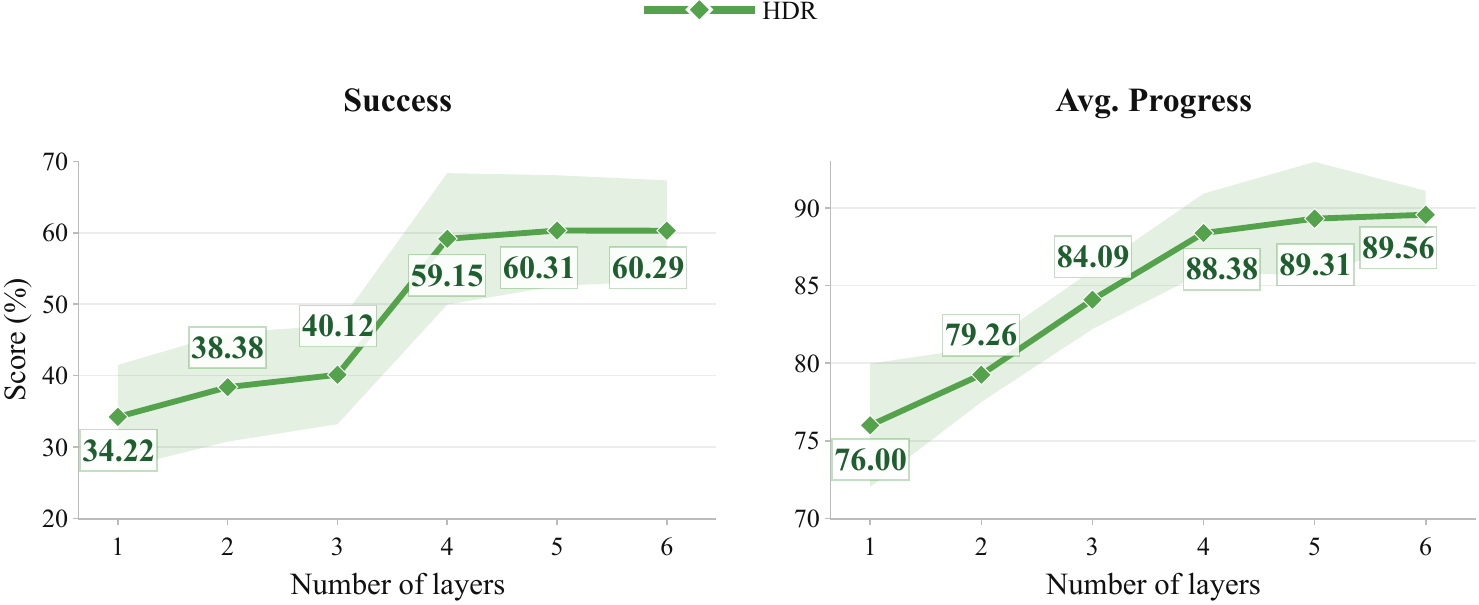}
\end{minipage}
\hfill
\begin{minipage}[c]{0.38\textwidth}
    \caption{Hierarchical layer importance. Curves show mean performance with one-standard-deviation bands. Increasing active hierarchy layers from 1 to 6 consistently improves HDR over CausalForcing~\cite{zhu2026causalforcingautoregressivediffusion}, showing that each layer contributes to the full coarse-to-fine reasoning process.}
    \label{fig:layer_reduction}
\end{minipage}
\vspace{-10pt}
\end{figure*}

\noindent\textbf{Hierarchical layer importance.}
We study how performance changes as the number of active hierarchical layers increases. Figure~\ref{fig:layer_reduction} reorganizes the variants by layer count, from a single-layer setting equivalent to CausalForcing to the full six-layer HDR hierarchy. The one-layer setting lacks the coarse-to-fine reasoning process enabled by the hierarchy. As additional layers are introduced, the model gains progressively richer high-level planning and stronger multi-step reasoning behavior. This trend shows that HDR's advantage does not come only from the final frame-level refinement: each hierarchy level contributes useful structure, and deeper hierarchies lead to better performance.

\subsection{Robustness: Performance under Reduced Budgets and Limited Data}
\label{sec:robustness}

\begin{figure*}[t]
    \centering

    \begin{subfigure}[t]{0.49\textwidth}
        \vspace{0pt}
        \centering
        \includegraphics[width=\linewidth]{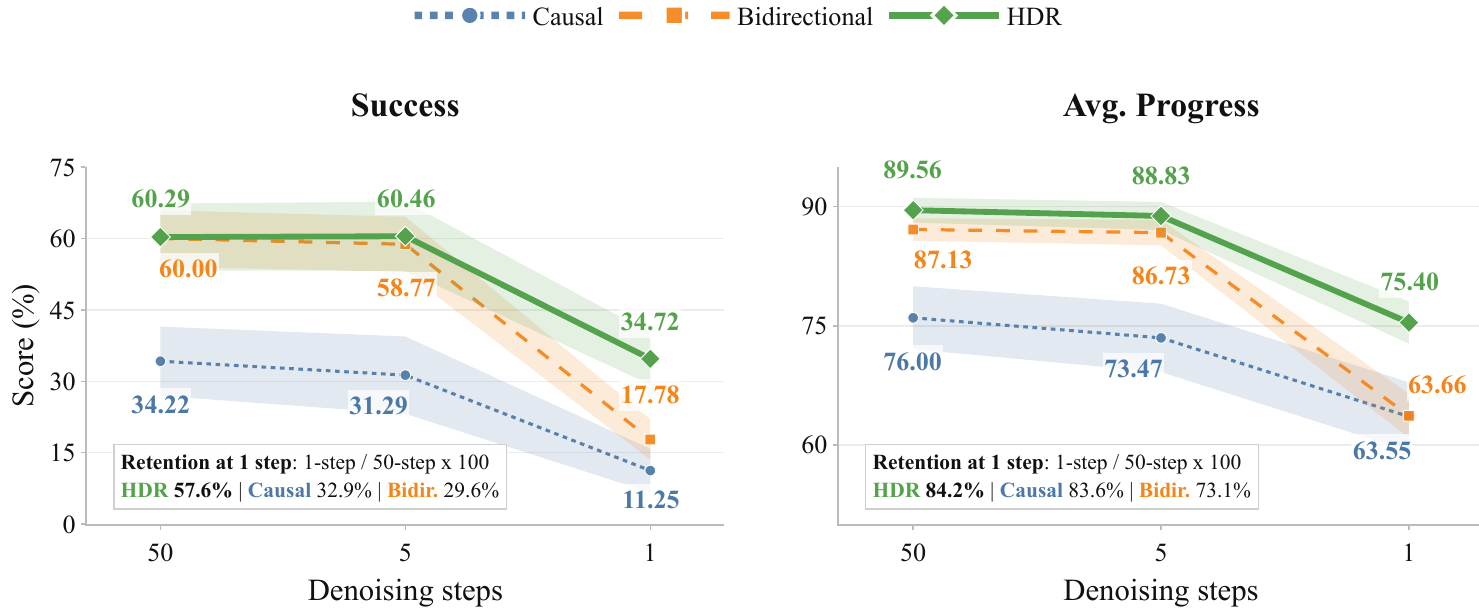}
        \caption{Denoising-step reduction. HDR remains more robust than both CausalForcing~\cite{zhu2026causalforcingautoregressivediffusion}, the streaming AR diffusion baseline, and bidirectional diffusion.}
        \label{fig:noisy_reduction}
    \end{subfigure}
    \hfill
    \begin{subfigure}[t]{0.49\textwidth}
        \vspace{0pt}
        \centering
        \includegraphics[width=\linewidth]{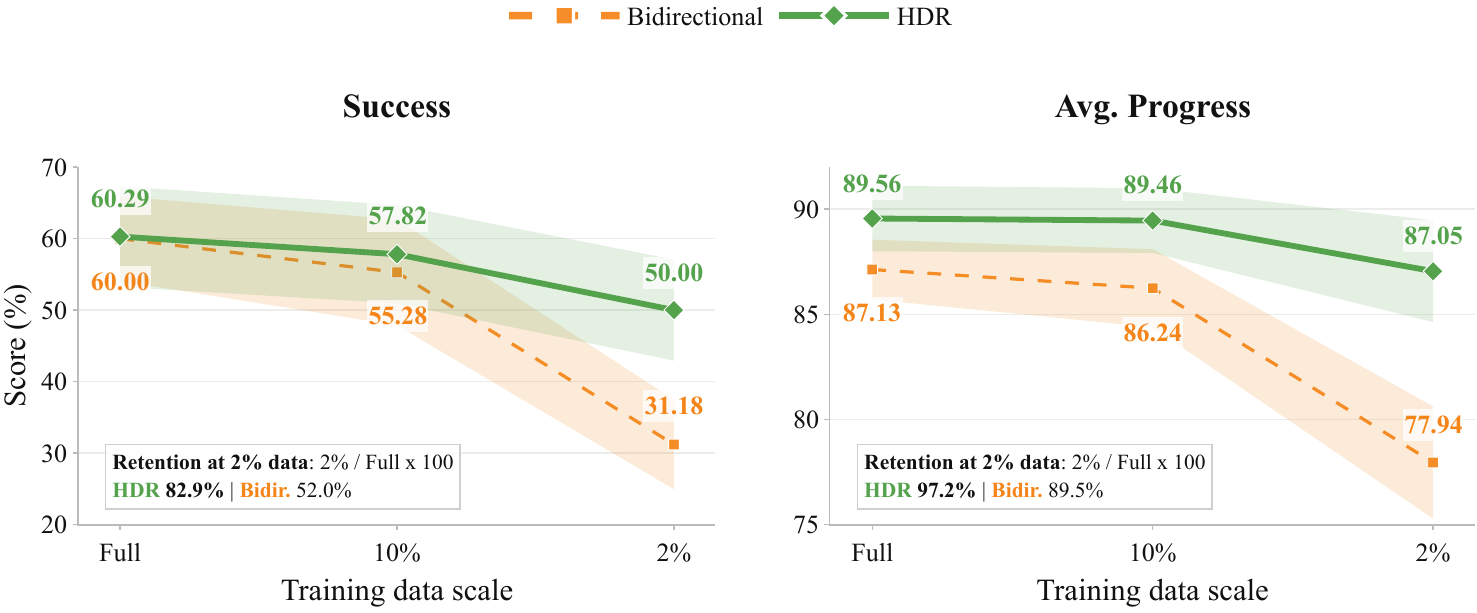}
        \caption{Data reduction. HDR degrades more gracefully than bidirectional diffusion as the training data scale decreases from the full set to 10\% and 2\%.}
        \label{fig:data_reduction}
    \end{subfigure}

    \caption{Robustness ablations of HDR. Curves show mean performance, and shaded bands denote one standard deviation. The left plot evaluates reduced denoising budgets, and the right plot evaluates reduced training-data scales. Full task-level data-reduction results are provided in Appendix Table~\ref{tab:data_ablation}.}
    \label{fig:efficiency_ablation}
    \vspace{-10pt}
\end{figure*}

\noindent\textbf{Denoising-step reduction.}
We first study robustness under reduced denoising budgets. Figure~\ref{fig:efficiency_ablation}(a) compares CausalForcing, bidirectional diffusion, and HDR as the number of inference denoising steps decreases. Both baselines degrade substantially under aggressive step reduction: bidirectional diffusion drops from 60.00 to 17.78 in overall success with one step, while CausalForcing drops from 34.22 to 11.25. In contrast, HDR remains substantially more robust, achieving 34.72 success with one denoising step and preserving 57.6\% of its full-step performance, compared with 29.6\% and 32.9\% for the bidirectional and CausalForcing baselines, respectively.

\noindent\textbf{Data reduction.}
We test whether HDR can learn reasoning rules from limited data. Figure~\ref{fig:efficiency_ablation}(b) compares HDR and bidirectional diffusion using the full training set, 10\%, and 2\% of the data, with task-level results in Appendix Table~\ref{tab:data_ablation}. As data decreases, HDR degrades more gracefully: with only 2\% of the training data, it retains 82.9\% of its full-data success and 97.2\% of its full-data average progress, compared with 52.0\% and 89.5\% for bidirectional diffusion. This suggests that the hierarchy provides a useful inductive bias for learning transferable task rules rather than simply memorizing training examples.

Together, these ablations show that HDR's robustness comes from its hierarchical architecture. The model remains effective under limited denoising budgets because coarse layers provide stable global guidance, while lower layers refine this guidance into detailed visual states. Conversely, when coarse levels are absent or training data is limited, the hierarchical reasoning process becomes the key factor that determines whether the model can preserve multi-step logical consistency.

\subsection{Physical World Modeling: Transfer to Robot Interaction}
\label{sec:physical_world}

To evaluate whether HDR transfers beyond synthetic benchmarks, we conduct a physical-world robot maze experiment. The task requires a robot arm to pick up a plush toy and move it through a maze built from toy blocks. This setting introduces visual domain shifts, imperfect object localization, occlusion, lighting variation, and irregular maze boundaries. We pretrain HDR on 3,000 virtual maze videos, fine-tune both HDR and CausalForcing using only 50 real-world robot videos, and convert generated videos into executable robot actions using an inverse dynamics model (IDM).

\begin{figure*}[t]
\centering
\begin{minipage}[c]{0.68\textwidth}
    \centering
    \includegraphics[width=\linewidth]{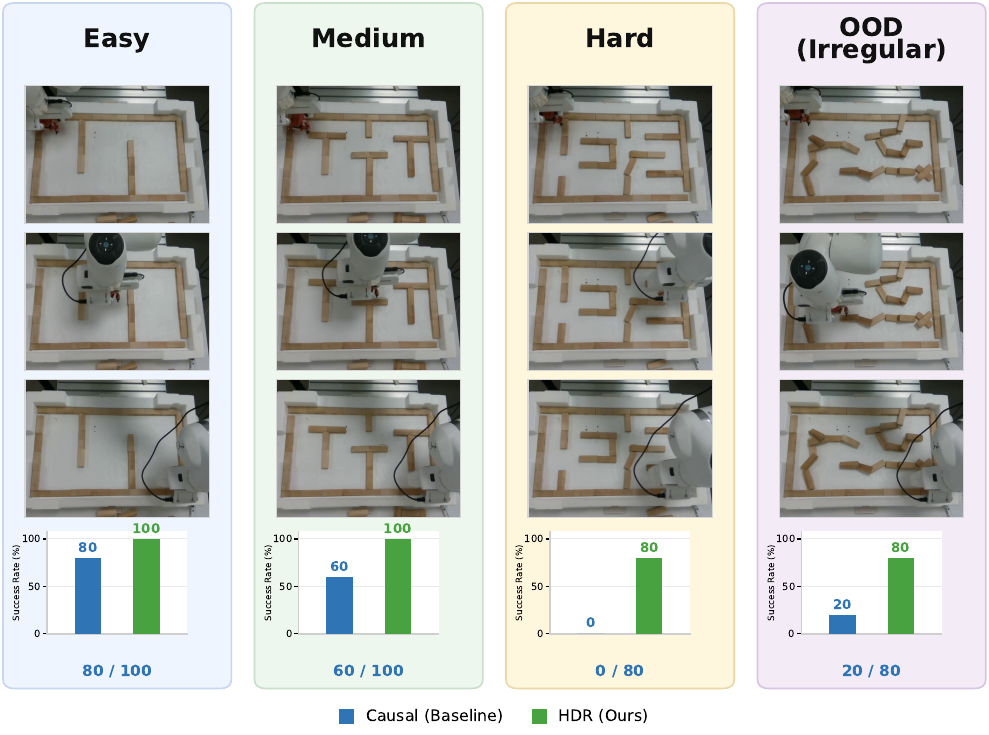}
\end{minipage}
\hfill
\begin{minipage}[c]{0.29\textwidth}
    \caption{Physical-world robot maze experiment. We fine-tune both CausalForcing~\cite{zhu2026causalforcingautoregressivediffusion} and HDR using only 50 real-world robot maze videos, then use an inverse dynamics model (IDM) to convert generated videos into executable robot actions. HDR maintains strong success rates across easy, medium, hard, and OOD mazes, where OOD mazes contain diagonal or stacked wooden blocks.}
    \label{fig:realworld_vis}
\end{minipage}
\vspace{-10pt}
\end{figure*}

We categorize physical-world mazes by difficulty. Easy, medium, and hard mazes are defined by maze complexity, including the number of branches and turns required by the solution path. We also include an OOD setting, where wooden blocks are placed diagonally or stacked together, producing layouts that differ from regular grid-like training mazes. As shown in Figure~\ref{fig:realworld_vis}, HDR maintains strong success rates across all settings, indicating that hierarchical denoising can transfer multi-step reasoning ability to physical interaction.

\subsection{World Action Modeling on RoboDojo}
\label{sec:hdr_wam_robodojo}

We also evaluate whether hierarchical denoising transfers to embodied world-action modeling. We instantiate \textbf{HDR-WAM} by combining episode-level visual context with a local action-conditioned rollout, while keeping the detailed token layout, sampling procedure, and attention mask in Appendix~\ref{sec:appendix_hdr_wam_details}.

Table~\ref{tab:robodojo_leaderboard} reports results on the RoboDojo simulation benchmark, which contains 42 robot interaction tasks grouped into five capability dimensions. Without robot-domain or embodied-interaction pretraining, HDR-WAM reaches an overall score of \(5.47\) and an average success rate of \(3.00\%\). Among no-pretraining World Action Model baselines, it improves over AHA-WAM (\(4.82 / 2.39\%\)) and Fast-WAM (\(3.48 / 2.03\%\)), establishing a new no-pretraining WAM state of the art on the benchmark.

\begin{figure*}[t]
\centering
\begin{minipage}[c]{0.68\textwidth}
    \centering
    \includegraphics[width=\linewidth]{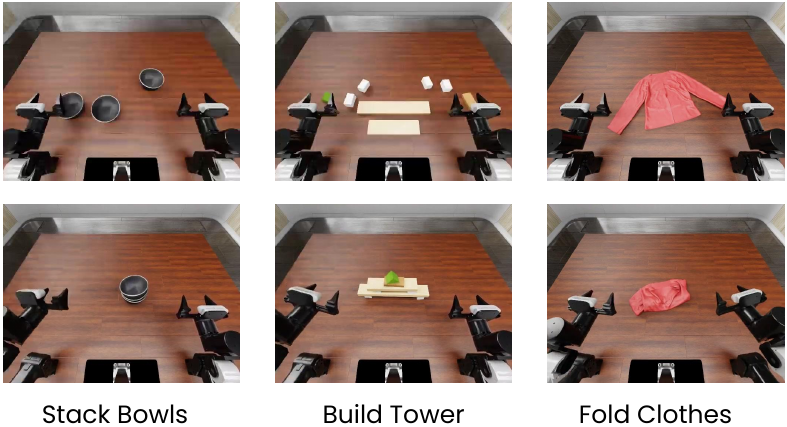}
\end{minipage}
\hfill
\begin{minipage}[c]{0.29\textwidth}
    \caption{RoboDojo environments and HDR-WAM executions. The figure presents representative tabletop manipulation environments from the RoboDojo simulation benchmark together with executions produced by HDR-WAM. The model achieves particularly strong performance on Long-Horizon tasks, reaching a score of \(9.85\) and a success rate of \(4.75\%\), demonstrating its ability to maintain coherent task structure over extended interactions.}
    \label{fig:robodojo}
\end{minipage}
\vspace{-10pt}
\end{figure*}

The strongest gains appear in the Long-Horizon and Memory dimensions, where HDR-WAM reaches \(9.85 / 4.75\%\) and \(6.65 / 4.67\%\), respectively. This pattern matches the intended role of hierarchical denoising: sparse episode-level landmarks help preserve coherent task structure over extended interactions, while the local action-conditioned rollout keeps the actor responsive to the current observation. Figure~\ref{fig:robodojo} visualizes representative RoboDojo environments and HDR-WAM executions, further illustrating the model's strong long-horizon task capability.

\begin{table*}[t]
\centering
\footnotesize
\caption{RoboDojo simulation benchmark leaderboard. Each cell reports score / success rate for a capability dimension. Within each pretraining group, the best and second-best scores in each column are shown in \textbf{bold} and \underline{underlined}, respectively, with ties included. The Pretrain column indicates whether the method uses robot-domain or embodied-interaction pretraining beyond task-specific benchmark training. HDR-WAM adapts \ours{} to World Action Models through hierarchical episode-conditioned visual planning and local action prediction.}
\label{tab:robodojo_leaderboard}
\setlength{\tabcolsep}{3.0pt}
\renewcommand{\arraystretch}{1.12}
\resizebox{\textwidth}{!}{%
\begin{tabular}{lccccccc}
\toprule
Alg.\textbackslash Dim 
& Pretrain
& Generalization 
& Precision 
& Long-Horizon 
& Memory 
& Open 
& Average \\
\midrule

X-WAM \cite{guo2026unified4dworldaction}
& Yes
& \textbf{7.39 / 3.33\%} 
& \textbf{6.72 / 1.83\%} 
& \textbf{17.47 / 9.08\%} 
& \textbf{6.32 / 4.67\%} 
& \underline{0.57 / 0.25\%} 
& \textbf{7.69 / 3.83\%} \\

GigaWorld-Policy \cite{ye2026gigaworldpolicyefficientactioncenteredworldaction}
& Yes
& 5.34 / 2.89\% 
& \underline{6.15 / 1.83\%} 
& \underline{15.51 / 8.92\%} 
& 3.46 / 2.22\% 
& 0.54 / 0.50\% 
& \underline{6.20 / 3.27\%} \\

LDA-1B \cite{lyu2026lda1bscalinglatentdynamics}
& Yes
& 0.71 / 0.17\% 
& 3.21 / 0.50\% 
& 1.92 / 0.08\% 
& 2.08 / 1.78\% 
& 0.00 / 0.00\% 
& 1.58 / 0.51\% \\

RDT-1B  \cite{liu2025rdt1bdiffusionfoundationmodel}
& Yes
& 0.56 / 0.33\% 
& 0.38 / 0.00\% 
& 1.13 / 0.00\% 
& 0.49 / 0.33\% 
& 0.00 / 0.00\% 
& 0.51 / 0.13\% \\

H-RDT \cite{bi2025hrdthumanmanipulationenhanced}
& Yes
& 0.49 / 0.22\% 
& 0.41 / 0.00\% 
& 2.23 / 0.17\% 
& 0.12 / 0.11\% 
& 0.08 / 0.08\% 
& 0.67 / 0.12\% \\

\midrule

\rowcolor{rowours}
\textbf{HDR-WAM (Ours)}
& \textbf{No}
& \underline{4.98 / 3.17\%}
& \textbf{5.90 / 2.25\%}
& \textbf{9.85 / 4.75\%}
& \textbf{6.65 / 4.67\%}
& \underline{0.50 / 0.50\%}
& \textbf{5.47 / 3.00\%} \\

AHA-WAM \cite{cai2026ahawamasynchronoushorizonadaptiveworldactionmodeling}
& No
& \textbf{5.79 / 3.28\%} 
& \underline{5.86 / 2.42\%} 
& 8.61 / 2.67\% 
& 2.97 / 2.78\% 
& \textbf{0.88 / 0.83\%} 
& \underline{4.82 / 2.39\%} \\

Fast-WAM \cite{yuan2026fastwamworldactionmodels}
& No
& 2.34 / 1.11\% 
& 1.96 / 0.00\% 
& \underline{9.14 / 5.17\%} 
& \underline{3.55 / 3.44\%} 
& 0.42 / 0.42\% 
& 3.48 / 2.03\% \\

ACT \cite{zhao2023learningfinegrainedbimanualmanipulation}
& No
& 0.69 / 0.56\% 
& 0.85 / 0.00\% 
& 1.73 / 0.92\% 
& 1.65 / 0.13\% 
& 0.00 / 0.00\% 
& 0.98 / 0.32\% \\

\bottomrule
\end{tabular}}
\vspace{-4pt}
\end{table*}

\section{Conclusion}

We introduced \ours{}, a framework for long-horizon video reasoning. By organizing video latents into a coarse-to-fine temporal tree, equipping the hierarchy with sparse structured attention, and allocating denoising budgets according to an entropy-matched principle, the method recovers much of the reasoning ability of bidirectional diffusion without relying on full temporal attention.

Empirically, \ours{} outperforms the causal baseline and remains competitive with bidirectional diffusion across six benchmarks. Ablations show both ingredients matter: removing coarse hierarchy levels hurts, and fully denoising every level is inferior to preserving uncertainty at upper levels.

More broadly, our results suggest that global reasoning in generative video modeling does not necessarily require dense all-to-all temporal computation. Structured multi-scale latent planning provides a promising alternative that preserves causal efficiency while preserving reasoning ability.

\bibliographystyle{plain}
\bibliography{references}

@misc{tong2022videomaemaskedautoencodersdataefficient,
      title={VideoMAE: Masked Autoencoders are Data-Efficient Learners for Self-Supervised Video Pre-Training}, 
      author={Zhan Tong and Yibing Song and Jue Wang and Limin Wang},
      year={2022},
      eprint={2203.12602},
      archivePrefix={arXiv},
      primaryClass={cs.CV},
      url={https://arxiv.org/abs/2203.12602}, 
}

@misc{zhang2025worldmodelsbenefitvlms,
            title={Can World Models Benefit VLMs for World Dynamics?},
            author={Kevin Zhang and Kuangzhi Ge and Xiaowei Chi and Renrui Zhang and Shaojun Shi and Zhen Dong and Sirui Han and
            Shanghang Zhang},
            year={2025},
            eprint={2510.00855},
            archivePrefix={arXiv},
            primaryClass={cs.CV},
            url={https://arxiv.org/abs/2510.00855},
            }

@misc{yan2021videogptvideogenerationusing,
      title={VideoGPT: Video Generation using VQ-VAE and Transformers}, 
      author={Wilson Yan and Yunzhi Zhang and Pieter Abbeel and Aravind Srinivas},
      year={2021},
      eprint={2104.10157},
      archivePrefix={arXiv},
      primaryClass={cs.CV},
      url={https://arxiv.org/abs/2104.10157}, 
}

@misc{zhu2026causalforcingautoregressivediffusion,
      title={Causal Forcing: Autoregressive Diffusion Distillation Done Right for High-Quality Real-Time Interactive Video Generation}, 
      author={Hongzhou Zhu and Min Zhao and Guande He and Hang Su and Chongxuan Li and Jun Zhu},
      year={2026},
      eprint={2602.02214},
      archivePrefix={arXiv},
      primaryClass={cs.CV},
      url={https://arxiv.org/abs/2602.02214}, 
}

@misc{wan2025wanopenadvancedlargescale,
      title={Wan: Open and Advanced Large-Scale Video Generative Models}, 
      author={Team Wan and Ang Wang and Baole Ai and Bin Wen and Chaojie Mao and Chen-Wei Xie and Di Chen and Feiwu Yu and Haiming Zhao and Jianxiao Yang and Jianyuan Zeng and Jiayu Wang and Jingfeng Zhang and Jingren Zhou and Jinkai Wang and Jixuan Chen and Kai Zhu and Kang Zhao and Keyu Yan and Lianghua Huang and Mengyang Feng and Ningyi Zhang and Pandeng Li and Pingyu Wu and Ruihang Chu and Ruili Feng and Shiwei Zhang and Siyang Sun and Tao Fang and Tianxing Wang and Tianyi Gui and Tingyu Weng and Tong Shen and Wei Lin and Wei Wang and Wei Wang and Wenmeng Zhou and Wente Wang and Wenting Shen and Wenyuan Yu and Xianzhong Shi and Xiaoming Huang and Xin Xu and Yan Kou and Yangyu Lv and Yifei Li and Yijing Liu and Yiming Wang and Yingya Zhang and Yitong Huang and Yong Li and You Wu and Yu Liu and Yulin Pan and Yun Zheng and Yuntao Hong and Yupeng Shi and Yutong Feng and Zeyinzi Jiang and Zhen Han and Zhi-Fan Wu and Ziyu Liu},
      year={2025},
      eprint={2503.20314},
      archivePrefix={arXiv},
      primaryClass={cs.CV},
      url={https://arxiv.org/abs/2503.20314}, 
}

@misc{blattmann2023stablevideodiffusionscaling,
      title={Stable Video Diffusion: Scaling Latent Video Diffusion Models to Large Datasets}, 
      author={Andreas Blattmann and Tim Dockhorn and Sumith Kulal and Daniel Mendelevitch and Maciej Kilian and Dominik Lorenz and Yam Levi and Zion English and Vikram Voleti and Adam Letts and Varun Jampani and Robin Rombach},
      year={2023},
      eprint={2311.15127},
      archivePrefix={arXiv},
      primaryClass={cs.CV},
      url={https://arxiv.org/abs/2311.15127}, 
}

@misc{nvidia2025cosmosworldfoundationmodel,
      title={Cosmos World Foundation Model Platform for Physical AI}, 
      author={NVIDIA and : and Niket Agarwal and Arslan Ali and Maciej Bala and Yogesh Balaji and Erik Barker and Tiffany Cai and Prithvijit Chattopadhyay and Yongxin Chen and Yin Cui and Yifan Ding and Daniel Dworakowski and Jiaojiao Fan and Michele Fenzi and Francesco Ferroni and Sanja Fidler and Dieter Fox and Songwei Ge and Yunhao Ge and Jinwei Gu and Siddharth Gururani and Ethan He and Jiahui Huang and Jacob Huffman and Pooya Jannaty and Jingyi Jin and Seung Wook Kim and Gergely Klár and Grace Lam and Shiyi Lan and Laura Leal-Taixe and Anqi Li and Zhaoshuo Li and Chen-Hsuan Lin and Tsung-Yi Lin and Huan Ling and Ming-Yu Liu and Xian Liu and Alice Luo and Qianli Ma and Hanzi Mao and Kaichun Mo and Arsalan Mousavian and Seungjun Nah and Sriharsha Niverty and David Page and Despoina Paschalidou and Zeeshan Patel and Lindsey Pavao and Morteza Ramezanali and Fitsum Reda and Xiaowei Ren and Vasanth Rao Naik Sabavat and Ed Schmerling and Stella Shi and Bartosz Stefaniak and Shitao Tang and Lyne Tchapmi and Przemek Tredak and Wei-Cheng Tseng and Jibin Varghese and Hao Wang and Haoxiang Wang and Heng Wang and Ting-Chun Wang and Fangyin Wei and Xinyue Wei and Jay Zhangjie Wu and Jiashu Xu and Wei Yang and Lin Yen-Chen and Xiaohui Zeng and Yu Zeng and Jing Zhang and Qinsheng Zhang and Yuxuan Zhang and Qingqing Zhao and Artur Zolkowski},
      year={2025},
      eprint={2501.03575},
      archivePrefix={arXiv},
      primaryClass={cs.CV},
      url={https://arxiv.org/abs/2501.03575}, 
}

@misc{yin2025slowbidirectionalfastautoregressive,
      title={From Slow Bidirectional to Fast Autoregressive Video Diffusion Models}, 
      author={Tianwei Yin and Qiang Zhang and Richard Zhang and William T. Freeman and Fredo Durand and Eli Shechtman and Xun Huang},
      year={2025},
      eprint={2412.07772},
      archivePrefix={arXiv},
      primaryClass={cs.CV},
      url={https://arxiv.org/abs/2412.07772}, 
}

@misc{huang2025selfforcingbridgingtraintest,
      title={Self Forcing: Bridging the Train-Test Gap in Autoregressive Video Diffusion}, 
      author={Xun Huang and Zhengqi Li and Guande He and Mingyuan Zhou and Eli Shechtman},
      year={2025},
      eprint={2506.08009},
      archivePrefix={arXiv},
      primaryClass={cs.CV},
      url={https://arxiv.org/abs/2506.08009}, 
}

@misc{robbyantteam2026advancingopensourceworldmodels,
      title={Advancing Open-source World Models}, 
      author={Robbyant Team and Zelin Gao and Qiuyu Wang and Yanhong Zeng and Jiapeng Zhu and Ka Leong Cheng and Yixuan Li and Hanlin Wang and Yinghao Xu and Shuailei Ma and Yihang Chen and Jie Liu and Yansong Cheng and Yao Yao and Jiayi Zhu and Yihao Meng and Kecheng Zheng and Qingyan Bai and Jingye Chen and Zehong Shen and Yue Yu and Xing Zhu and Yujun Shen and Hao Ouyang},
      year={2026},
      eprint={2601.20540},
      archivePrefix={arXiv},
      primaryClass={cs.CV},
      url={https://arxiv.org/abs/2601.20540}, 
}

@misc{wang2026bigvideoreasoningsuite,
      title={A Very Big Video Reasoning Suite}, 
      author={Maijunxian Wang and Ruisi Wang and Juyi Lin and Ran Ji and Thaddäus Wiedemer and Qingying Gao and Dezhi Luo and Yaoyao Qian and Lianyu Huang and Zelong Hong and Jiahui Ge and Qianli Ma and Hang He and Yifan Zhou and Lingzi Guo and Lantao Mei and Jiachen Li and Hanwen Xing and Tianqi Zhao and Fengyuan Yu and Weihang Xiao and Yizheng Jiao and Jianheng Hou and Danyang Zhang and Pengcheng Xu and Boyang Zhong and Zehong Zhao and Gaoyun Fang and John Kitaoka and Yile Xu and Hua Xu and Kenton Blacutt and Tin Nguyen and Siyuan Song and Haoran Sun and Shaoyue Wen and Linyang He and Runming Wang and Yanzhi Wang and Mengyue Yang and Ziqiao Ma and Raphaël Millière and Freda Shi and Nuno Vasconcelos and Daniel Khashabi and Alan Yuille and Yilun Du and Ziming Liu and Bo Li and Dahua Lin and Ziwei Liu and Vikash Kumar and Yijiang Li and Lei Yang and Zhongang Cai and Hokin Deng},
      year={2026},
      eprint={2602.20159},
      archivePrefix={arXiv},
      primaryClass={cs.CV},
      url={https://arxiv.org/abs/2602.20159}, 
}

@misc{wang2026demystifingvideoreasoning,
      title={Demystifing Video Reasoning}, 
      author={Ruisi Wang and Zhongang Cai and Fanyi Pu and Junxiang Xu and Wanqi Yin and Maijunxian Wang and Ran Ji and Chenyang Gu and Bo Li and Ziqi Huang and Hokin Deng and Dahua Lin and Ziwei Liu and Lei Yang},
      year={2026},
      eprint={2603.16870},
      archivePrefix={arXiv},
      primaryClass={cs.CV},
      url={https://arxiv.org/abs/2603.16870}, 
}

@misc{wu2026visualgenerationunlockshumanlike,
      title={Visual Generation Unlocks Human-Like Reasoning through Multimodal World Models}, 
      author={Jialong Wu and Xiaoying Zhang and Hongyi Yuan and Xiangcheng Zhang and Tianhao Huang and Changjing He and Chaoyi Deng and Renrui Zhang and Youbin Wu and Mingsheng Long},
      year={2026},
      eprint={2601.19834},
      archivePrefix={arXiv},
      primaryClass={cs.AI},
      url={https://arxiv.org/abs/2601.19834}, 
}

@misc{luo2025vreasonbenchunifiedreasoningbenchmark,
      title={V-ReasonBench: Toward Unified Reasoning Benchmark Suite for Video Generation Models}, 
      author={Yang Luo and Xuanlei Zhao and Baijiong Lin and Lingting Zhu and Liyao Tang and Yuqi Liu and Ying-Cong Chen and Shengju Qian and Xin Wang and Yang You},
      year={2025},
      eprint={2511.16668},
      archivePrefix={arXiv},
      primaryClass={cs.CV},
      url={https://arxiv.org/abs/2511.16668}, 
}

@misc{yu2025vrbenchbenchmarkmultistepreasoning,
      title={VRBench: A Benchmark for Multi-Step Reasoning in Long Narrative Videos}, 
      author={Jiashuo Yu and Yue Wu and Meng Chu and Zhifei Ren and Zizheng Huang and Pei Chu and Ruijie Zhang and Yinan He and Qirui Li and Songze Li and Zhenxiang Li and Zhongying Tu and Conghui He and Yu Qiao and Yali Wang and Yi Wang and Limin Wang},
      year={2025},
      eprint={2506.10857},
      archivePrefix={arXiv},
      primaryClass={cs.CV},
      url={https://arxiv.org/abs/2506.10857}, 
}

@misc{henschel2025streamingt2vconsistentdynamicextendable,
      title={StreamingT2V: Consistent, Dynamic, and Extendable Long Video Generation from Text}, 
      author={Roberto Henschel and Levon Khachatryan and Hayk Poghosyan and Daniil Hayrapetyan and Vahram Tadevosyan and Zhangyang Wang and Shant Navasardyan and Humphrey Shi},
      year={2025},
      eprint={2403.14773},
      archivePrefix={arXiv},
      primaryClass={cs.CV},
      url={https://arxiv.org/abs/2403.14773}, 
}

@misc{kim2024fifodiffusiongeneratinginfinitevideos,
      title={FIFO-Diffusion: Generating Infinite Videos from Text without Training}, 
      author={Jihwan Kim and Junoh Kang and Jinyoung Choi and Bohyung Han},
      year={2024},
      eprint={2405.11473},
      archivePrefix={arXiv},
      primaryClass={cs.CV},
      url={https://arxiv.org/abs/2405.11473}, 
}

@misc{li2025arlonboostingdiffusiontransformers,
      title={ARLON: Boosting Diffusion Transformers with Autoregressive Models for Long Video Generation}, 
      author={Zongyi Li and Shujie Hu and Shujie Liu and Long Zhou and Jeongsoo Choi and Lingwei Meng and Xun Guo and Jinyu Li and Hefei Ling and Furu Wei},
      year={2025},
      eprint={2410.20502},
      archivePrefix={arXiv},
      primaryClass={cs.CV},
      url={https://arxiv.org/abs/2410.20502}, 
}

@misc{gao2025ca2vdmefficientautoregressivevideo,
      title={Ca2-VDM: Efficient Autoregressive Video Diffusion Model with Causal Generation and Cache Sharing}, 
      author={Kaifeng Gao and Jiaxin Shi and Hanwang Zhang and Chunping Wang and Jun Xiao and Long Chen},
      year={2025},
      eprint={2411.16375},
      archivePrefix={arXiv},
      primaryClass={cs.CV},
      url={https://arxiv.org/abs/2411.16375}, 
}

@misc{yang2025longliverealtimeinteractivelong,
      title={LongLive: Real-time Interactive Long Video Generation}, 
      author={Shuai Yang and Wei Huang and Ruihang Chu and Yicheng Xiao and Yuyang Zhao and Xianbang Wang and Muyang Li and Enze Xie and Yingcong Chen and Yao Lu and Song Han and Yukang Chen},
      year={2025},
      eprint={2509.22622},
      archivePrefix={arXiv},
      primaryClass={cs.CV},
      url={https://arxiv.org/abs/2509.22622}, 
}

@misc{liu2025rollingforcingautoregressivelong,
      title={Rolling Forcing: Autoregressive Long Video Diffusion in Real Time}, 
      author={Kunhao Liu and Wenbo Hu and Jiale Xu and Ying Shan and Shijian Lu},
      year={2025},
      eprint={2509.25161},
      archivePrefix={arXiv},
      primaryClass={cs.CV},
      url={https://arxiv.org/abs/2509.25161}, 
}

@misc{deng2025autoregressivevideogenerationvector,
      title={Autoregressive Video Generation without Vector Quantization}, 
      author={Haoge Deng and Ting Pan and Haiwen Diao and Zhengxiong Luo and Yufeng Cui and Huchuan Lu and Shiguang Shan and Yonggang Qi and Xinlong Wang},
      year={2025},
      eprint={2412.14169},
      archivePrefix={arXiv},
      primaryClass={cs.CV},
      url={https://arxiv.org/abs/2412.14169}, 
}

@misc{wiedemer2025videomodelszeroshotlearners,
      title={Video models are zero-shot learners and reasoners}, 
      author={Thaddäus Wiedemer and Yuxuan Li and Paul Vicol and Shixiang Shane Gu and Nick Matarese and Kevin Swersky and Been Kim and Priyank Jaini and Robert Geirhos},
      year={2025},
      eprint={2509.20328},
      archivePrefix={arXiv},
      primaryClass={cs.LG},
      url={https://arxiv.org/abs/2509.20328}, 
}

@misc{qian2025wristworldgeneratingwristviews4d,
      title={WristWorld: Generating Wrist-Views via 4D World Models for Robotic Manipulation}, 
      author={Zezhong Qian and Xiaowei Chi and Yuming Li and Shizun Wang and Zhiyuan Qin and Xiaozhu Ju and Sirui Han and Shanghang Zhang},
      year={2025},
      eprint={2510.07313},
      archivePrefix={arXiv},
      primaryClass={cs.CV},
      url={https://arxiv.org/abs/2510.07313}, 
}

@inproceedings{
pyramid,
title={Pyramidal Flow Matching for Efficient Video Generative Modeling},
author={Yang Jin and Zhicheng Sun and Ningyuan Li and Kun Xu and Kun Xu and Hao Jiang and Nan Zhuang and Quzhe Huang and Yang Song and Yadong MU and Zhouchen Lin},
booktitle={The Thirteenth International Conference on Learning Representations},
year={2025}
}

@inproceedings{
xiao2024efficient,
title={Efficient Streaming Language Models with Attention Sinks},
author={Guangxuan Xiao and Yuandong Tian and Beidi Chen and Song Han and Mike Lewis},
booktitle={The Twelfth International Conference on Learning Representations},
year={2024}
}

@inproceedings{jia2025ditar,
  title={DiTAR: Diffusion Transformer Autoregressive Modeling for Speech Generation},
  author={Jia, Dongya and Chen, Zhuo and Chen, Jiawei and Du, Chenpeng and Wu, Jian and Cong, Jian and Zhuang, Xiaobin and Li, Chumin and Wei, Zhen and Wang, Yuping and others},
  booktitle={International Conference on Machine Learning},
  pages={27255--27270},
  year={2025},
  organization={PMLR}
}

@misc{li2026causalworldmodelingrobot,
      title={Causal World Modeling for Robot Control}, 
      author={Lin Li and Qihang Zhang and Yiming Luo and Shuai Yang and Ruilin Wang and Fei Han and Mingrui Yu and Zelin Gao and Nan Xue and Xing Zhu and Yujun Shen and Yinghao Xu},
      year={2026},
      eprint={2601.21998},
      archivePrefix={arXiv},
      primaryClass={cs.CV},
      url={https://arxiv.org/abs/2601.21998}, 
}

@misc{yu2025contextmemorysceneconsistentinteractive,
      title={Context as Memory: Scene-Consistent Interactive Long Video Generation with Memory Retrieval}, 
      author={Jiwen Yu and Jianhong Bai and Yiran Qin and Quande Liu and Xintao Wang and Pengfei Wan and Di Zhang and Xihui Liu},
      year={2025},
      eprint={2506.03141},
      archivePrefix={arXiv},
      primaryClass={cs.CV},
      url={https://arxiv.org/abs/2506.03141}, 
}

@misc{ye2026worldactionmodelszeroshot,
      title={World Action Models are Zero-shot Policies}, 
      author={Seonghyeon Ye and Yunhao Ge and Kaiyuan Zheng and Shenyuan Gao and Sihyun Yu and George Kurian and Suneel Indupuru and You Liang Tan and Chuning Zhu and Jiannan Xiang and Ayaan Malik and Kyungmin Lee and William Liang and Nadun Ranawaka and Jiasheng Gu and Yinzhen Xu and Guanzhi Wang and Fengyuan Hu and Avnish Narayan and Johan Bjorck and Jing Wang and Gwanghyun Kim and Dantong Niu and Ruijie Zheng and Yuqi Xie and Jimmy Wu and Qi Wang and Ryan Julian and Danfei Xu and Yilun Du and Yevgen Chebotar and Scott Reed and Jan Kautz and Yuke Zhu and Linxi "Jim" Fan and Joel Jang},
      year={2026},
      eprint={2602.15922},
      archivePrefix={arXiv},
      primaryClass={cs.RO},
      url={https://arxiv.org/abs/2602.15922}, 
}

@misc{hong2025relicinteractivevideoworld,
      title={RELIC: Interactive Video World Model with Long-Horizon Memory}, 
      author={Yicong Hong and Yiqun Mei and Chongjian Ge and Yiran Xu and Yang Zhou and Sai Bi and Yannick Hold-Geoffroy and Mike Roberts and Matthew Fisher and Eli Shechtman and Kalyan Sunkavalli and Feng Liu and Zhengqi Li and Hao Tan},
      year={2025},
      eprint={2512.04040},
      archivePrefix={arXiv},
      primaryClass={cs.CV},
      url={https://arxiv.org/abs/2512.04040}, 
}

@misc{yu2024languagemodelbeatsdiffusion,
      title={Language Model Beats Diffusion -- Tokenizer is Key to Visual Generation}, 
      author={Lijun Yu and José Lezama and Nitesh B. Gundavarapu and Luca Versari and Kihyuk Sohn and David Minnen and Yong Cheng and Vighnesh Birodkar and Agrim Gupta and Xiuye Gu and Alexander G. Hauptmann and Boqing Gong and Ming-Hsuan Yang and Irfan Essa and David A. Ross and Lu Jiang},
      year={2024},
      eprint={2310.05737},
      archivePrefix={arXiv},
      primaryClass={cs.CV},
      url={https://arxiv.org/abs/2310.05737}, 
}

@misc{kondratyuk2024videopoetlargelanguagemodel,
      title={VideoPoet: A Large Language Model for Zero-Shot Video Generation}, 
      author={Dan Kondratyuk and Lijun Yu and Xiuye Gu and José Lezama and Jonathan Huang and Grant Schindler and Rachel Hornung and Vighnesh Birodkar and Jimmy Yan and Ming-Chang Chiu and Krishna Somandepalli and Hassan Akbari and Yair Alon and Yong Cheng and Josh Dillon and Agrim Gupta and Meera Hahn and Anja Hauth and David Hendon and Alonso Martinez and David Minnen and Mikhail Sirotenko and Kihyuk Sohn and Xuan Yang and Hartwig Adam and Ming-Hsuan Yang and Irfan Essa and Huisheng Wang and David A. Ross and Bryan Seybold and Lu Jiang},
      year={2024},
      eprint={2312.14125},
      archivePrefix={arXiv},
      primaryClass={cs.CV},
      url={https://arxiv.org/abs/2312.14125}, 
}

@misc{zhao2023learningfinegrainedbimanualmanipulation,
      title={Learning Fine-Grained Bimanual Manipulation with Low-Cost Hardware}, 
      author={Tony Z. Zhao and Vikash Kumar and Sergey Levine and Chelsea Finn},
      year={2023},
      eprint={2304.13705},
      archivePrefix={arXiv},
      primaryClass={cs.RO},
      url={https://arxiv.org/abs/2304.13705}, 
}

@misc{yuan2026fastwamworldactionmodels,
      title={Fast-WAM: Do World Action Models Need Test-time Future Imagination?}, 
      author={Tianyuan Yuan and Zibin Dong and Yicheng Liu and Hang Zhao},
      year={2026},
      eprint={2603.16666},
      archivePrefix={arXiv},
      primaryClass={cs.CV},
      url={https://arxiv.org/abs/2603.16666}, 
}

@misc{cai2026ahawamasynchronoushorizonadaptiveworldactionmodeling,
      title={AHA-WAM:Asynchronous Horizon-Adaptive World-Action Modeling with Observation-Guided Context Routing}, 
      author={Jisong Cai and Long Ling and Shiwei Chu and Zhongshan Liu and Jiayue Kang and Zhixuan Liang and Wenjie Xu and Yinan Mao and Weinan Zhang and Xiaokang Yang and Ru Ying and Ran Zheng and Yao Mu},
      year={2026},
      eprint={2606.09811},
      archivePrefix={arXiv},
      primaryClass={cs.RO},
      url={https://arxiv.org/abs/2606.09811}, 
}

@misc{guo2026unified4dworldaction,
      title={Unified 4D World Action Modeling from Video Priors with Asynchronous Denoising}, 
      author={Jun Guo and Qiwei Li and Peiyan Li and Zilong Chen and Nan Sun and Yifei Su and Heyun Wang and Yuan Zhang and Xinghang Li and Huaping Liu},
      year={2026},
      eprint={2604.26694},
      archivePrefix={arXiv},
      primaryClass={cs.RO},
      url={https://arxiv.org/abs/2604.26694}, 
}

@misc{ye2026gigaworldpolicyefficientactioncenteredworldaction,
      title={GigaWorld-Policy: An Efficient Action-Centered World--Action Model}, 
      author={Angen Ye and Boyuan Wang and Chaojun Ni and Guan Huang and Guosheng Zhao and Hao Li and Hengtao Li and Jie Li and Jindi Lv and Jingyu Liu and Min Cao and Peng Li and Qiuping Deng and Wenjun Mei and Xiaofeng Wang and Xinze Chen and Xinyu Zhou and Yang Wang and Yifan Chang and Yifan Li and Yukun Zhou and Yun Ye and Zhichao Liu and Zheng Zhu},
      year={2026},
      eprint={2603.17240},
      archivePrefix={arXiv},
      primaryClass={cs.CV},
      url={https://arxiv.org/abs/2603.17240}, 
}

@misc{lyu2026lda1bscalinglatentdynamics,
      title={LDA-1B: Scaling Latent Dynamics Action Model via Universal Embodied Data Ingestion}, 
      author={Jiangran Lyu and Kai Liu and Xuheng Zhang and Haoran Liao and Yusen Feng and Wenxuan Zhu and Tingrui Shen and Jiayi Chen and Jiazhao Zhang and Yifei Dong and Wenbo Cui and Senmao Qi and Shuo Wang and Yixin Zheng and Mi Yan and Xuesong Shi and Haoran Li and Dongbin Zhao and Ming-Yu Liu and Zhizheng Zhang and Li Yi and Yizhou Wang and He Wang},
      year={2026},
      eprint={2602.12215},
      archivePrefix={arXiv},
      primaryClass={cs.RO},
      url={https://arxiv.org/abs/2602.12215}, 
}

@misc{liu2025rdt1bdiffusionfoundationmodel,
      title={RDT-1B: a Diffusion Foundation Model for Bimanual Manipulation}, 
      author={Songming Liu and Lingxuan Wu and Bangguo Li and Hengkai Tan and Huayu Chen and Zhengyi Wang and Ke Xu and Hang Su and Jun Zhu},
      year={2025},
      eprint={2410.07864},
      archivePrefix={arXiv},
      primaryClass={cs.RO},
      url={https://arxiv.org/abs/2410.07864}, 
}

@misc{bi2025hrdthumanmanipulationenhanced,
      title={H-RDT: Human Manipulation Enhanced Bimanual Robotic Manipulation}, 
      author={Hongzhe Bi and Lingxuan Wu and Tianwei Lin and Hengkai Tan and Zhizhong Su and Hang Su and Jun Zhu},
      year={2025},
      eprint={2507.23523},
      archivePrefix={arXiv},
      primaryClass={cs.RO},
      url={https://arxiv.org/abs/2507.23523}, 
}

\newpage

\appendix

\section{HDR-WAM Details}
\label{sec:appendix_hdr_wam_details}

\subsection{Hierarchical Action Modeling}
\label{sec:appendix_hdr_wam_principle}

HDR-WAM adapts \ours{} to embodied world-action modeling by treating control as a joint denoising problem over visual dynamics and actions. The video stream predicts action-conditioned future observations, while the action stream predicts executable action chunks conditioned on language, proprioception, and visual context.

\paragraph{Temporal views.}
Consider an episode \(\mathcal{E}=\{(x_t,q_t,u_t)\}_{t=0}^{T-1}\), where \(x_t\) is the multi-view RGB observation, \(q_t\) is proprioception, and \(u_t\in\mathbb{R}^{d_a}\) is the low-level action. For a training sample starting at frame \(s\), HDR-WAM constructs two visual views. The episode-level view is a uniformly subsampled set of global anchors
\[
\mathcal{I}^{\mathrm{epi}}
=\left\{\left\lfloor \frac{i(T-1)}{N_e-1}\right\rceil \right\}_{i=0}^{N_e-1},
\qquad
X^{\mathrm{epi}}=\{x_i:i\in\mathcal{I}^{\mathrm{epi}}\},
\]
which summarizes task phase and long-range progress. The local action-conditioned view contains a local rollout window plus future visual landmarks. With local length \(N_\ell\), sampling stride \(\Delta\), and \(N_f\) future landmarks, we first take
\[
\mathcal{I}^{\mathrm{loc}}
=\left\{s+i\Delta\right\}_{i=0}^{N_\ell-1},
\qquad
t_{\mathrm{end}}=\min(s+(N_\ell-1)\Delta,T-1).
\]
The future landmark indices are then uniformly sampled after the local window,
\[
\mathcal{I}^{\mathrm{fut}}
=\left\{\left\lfloor t_{\mathrm{end}}+\frac{j}{N_f}(T-1-t_{\mathrm{end}})\right\rceil\right\}_{j=1}^{N_f},
\]
with the last available frame repeated when the remaining suffix is shorter than required. The local visual view is \(X^{\mathrm{loc}}=\{x_i:i\in\mathcal{I}^{\mathrm{loc}}\cup\mathcal{I}^{\mathrm{fut}}\}\). In our RoboDojo experiments, \(N_e=9\), \(N_\ell=9\), and \(N_f=4\).

\paragraph{Action alignment and token layout.}
Actions are aligned only to the \(N_\ell-1\) local visual transitions, not to the episode anchors or future landmarks. Given an action horizon \(H\), we split the action chunk into \(N_\ell-1\) groups,
\[
A_s=\{u_s,\ldots,u_{s+H-1}\},
\qquad
G_r=\{u_{s+r m},\ldots,u_{s+(r+1)m-1}\},
\qquad
m=\frac{H}{N_\ell-1},
\]
where \(r=0,\ldots,N_\ell-2\). Thus the additional global and future visual tokens provide context but do not introduce extra action targets. After VAE encoding, the two visual views produce latents \(Z^{\mathrm{epi}}\in\mathbb{R}^{C\times L_e\times H'\times W'}\) and \(Z^{\mathrm{loc}}\in\mathbb{R}^{C\times L_\ell\times H'\times W'}\). We concatenate them along time and keep the first latent of each view clean:
\[
Z=[Z^{\mathrm{epi}};Z^{\mathrm{loc}}],
\qquad
\mathcal{C}=\{0,L_e\}.
\]
For \(t\notin\mathcal{C}\), the training input is noised with the video diffusion process; for \(t\in\mathcal{C}\), the clean latent is preserved as an observed visual condition. The joint actor sequence is
\[
S=[V^{\mathrm{epi}},V^{\mathrm{loc}},A],
\]
where \(V^{\mathrm{epi}}\) and \(V^{\mathrm{loc}}\) are visual tokens obtained from the two latent views, and \(A\) denotes action tokens for the grouped action chunk.

\paragraph{Attention mask.}
HDR-WAM uses a block attention mask over the joint sequence \(S\) to separate visual denoising from action prediction. Let \(V=V^{\mathrm{epi}}\cup V^{\mathrm{loc}}\), let \(A\) be the action-token set, and let \(V_{\mathcal{C}}\subset V\) be tokens belonging to the clean visual latents. The MoT self-attention mask can be written as
\[
M_{\mathrm{MoT}}=
\begin{bmatrix}
M_{V\rightarrow V} & \mathbf{0}_{V\times A}\\
M_{A\rightarrow V} & \mathbf{1}_{A\times A}
\end{bmatrix},
\]
where \(M_{V\rightarrow V}\) is the video self-attention pattern, \(\mathbf{1}_{A\times A}\) allows action tokens to model the whole action chunk, and \(M_{A\rightarrow V}\) satisfies \(M_{A\rightarrow V}(a,v)=1\) only when \(v\in V_{\mathcal{C}}\). Video tokens do not directly attend to action tokens through MoT self-attention; instead, action conditioning enters the video denoiser through cross-attention to grouped action context. For the \(r\)-th local visual transition, the causal group mask is
\[
M^{\mathrm{cross}}_{r,k}=\mathbf{1}[k\le r],
\]
so denoising a visual transition can use the corresponding and previous action groups while the clean conditioning frames remain unconditioned by future actions. The video loss excludes clean indices \(\mathcal{C}\), and the action loss is computed on the action stream with padding masked out.

\begin{figure*}[t]
\centering
\includegraphics[width=\textwidth]{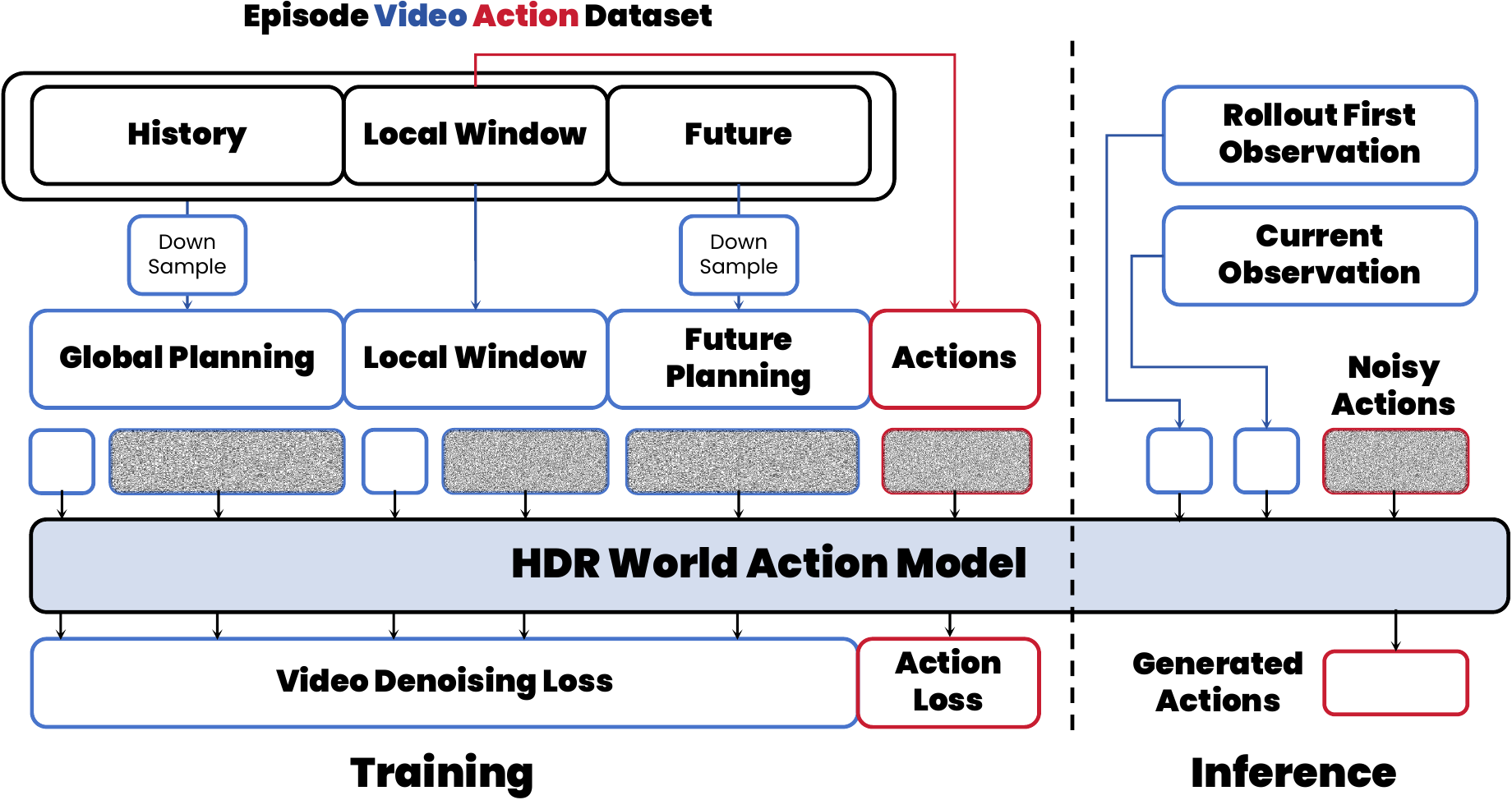}
\caption{Overview of the HDR-WAM actor. The model combines sparse episode-level visual context with a local action-conditioned rollout, then arranges episode, local visual, and action tokens in a joint sequence. The attention mask separates visual denoising from executable action prediction: action tokens attend to action tokens and clean visual conditions, while visual tokens denoise future observations under language, proprioception, and grouped action context.}
\label{fig:hdrwam_actor}
\vspace{-6pt}
\end{figure*}

\subsection{RoboDojo Data Processing}
\label{sec:appendix_hdr_wam_dataset}

For each training sample, we construct both temporal views from the same RoboDojo episode. The global view samples \(9\) RGB frames uniformly from the entire episode, forming sparse anchors for task-level progress. The local action-conditioned view starts at the current frame, takes \(9\) consecutive local RGB frames under the dataset stride, and appends \(4\) future landmarks sampled uniformly between the end of the local window and the end of the episode. If the remaining episode is shorter than required, the final available frame is repeated. Thus each local view contains \(13\) RGB frames, while action supervision remains aligned to the \(8\) transitions inside the local \(9\)-frame window.

RoboDojo observations use three camera views. We resize the top camera and the two wrist-side cameras, concatenate the side cameras horizontally, and stack the result under the top view before the standard resize, crop, and normalization pipeline. Both temporal views are encoded with the same video VAE and cached at the episode level, so the global anchors are shared by all samples from the same episode. This avoids repeatedly decoding the same long video while preserving the full episode context needed by HDR-WAM.

\section{Benchmark and Eval Details}
\label{sec:benchmark_eval_details}

We evaluate all methods on a mixed benchmark containing six long-horizon reasoning tasks: maze navigation, Tower of Hanoi, one-line drawing, sliding puzzle, Sokoban, and water pouring. These tasks cover diverse reasoning patterns, including spatial planning, state transition, object manipulation, and trajectory consistency.

Each generated video is evaluated with task-specific success criteria and reported using two metrics. The success score \(S\in\{0,1\}\) measures exact task completion under the benchmark-specific success criterion. The average progress score \(A\in[0,1]\) measures partial progress toward the target outcome and is therefore more tolerant of minor visual deviations that do not alter the core semantic trajectory. Results in the main paper are reported as Success / Avg.\ Progress pairs, with both values multiplied by \(100\).

For a scored set \(\mathcal{D}\), the reported task score is computed as the mean over evaluation samples:
\[
\text{Success}(\mathcal{D})=\frac{1}{|\mathcal{D}|}\sum_{i\in\mathcal{D}}S_i,
\qquad
\text{Avg.\ Progress}(\mathcal{D})=\frac{1}{|\mathcal{D}|}\sum_{i\in\mathcal{D}}A_i.
\]
The overall score is obtained by averaging the task-level scores across the six benchmarks. This gives equal weight to each reasoning task and reflects the model's robustness across different forms of long-horizon video reasoning.

The benchmark uses task-specific difficulty levels.  Maze and Hanoi are grouped by native problem size, while the remaining four tasks use level directories.  The evaluation split is summarized in Table~\ref{tab:appendix_levels}.

\begin{table}[h]
\centering
\small
\caption{Difficulty grouping of the mixed reasoning benchmark.}
\setlength{\tabcolsep}{1.5pt}
\begin{tabular}{l l l l}
\toprule
Task & Levels / Sizes & Eval Split & Main Difficulty Factor \\
\midrule
Maze & \(N=6,7,8,9,10\) & 10 per size & Grid size, path length \\
Hanoi & \(2,3,4,5\) disks & 10 ID + 10 OOD per size & Disk count, initial rods \\
One-line & levels \(2,3,4\) & 25 ID + 25 OOD per level & Path length, shape topology \\
Sliding & levels \(2,3,4\) & 30 ID + 20 OOD per level & Board size, optimal moves \\
Sokoban & levels \(2,3,4\) & 30 ID + 20 OOD per level & Layout grammar, action horizon \\
Water & levels \(2,3,4\) & 30 ID + 20 OOD per level & Tubes, capacity, solution length \\
\bottomrule
\end{tabular}
\label{tab:appendix_levels}
\end{table}

\subsection{Maze Navigation}

Maze samples are generated on square grids with fixed start \((0,0)\) and goal \((N-1,N-1)\).  The generator constructs a recursive-division maze graph, solves it with breadth-first search, stores the graph edges and solution path, and renders a red ball moving along the path.  The default evaluation set contains \(50\) samples, with \(10\) samples for each \(N\in\{6,7,8,9,10\}\).

The evaluator tracks the red ball using color segmentation and connected components.  The success score requires the decoded route to start correctly, remain in open cells, move only through valid graph edges, and reach the goal.  The average progress score is the normalized longest-common-subsequence overlap between the relaxed decoded route \(\hat{r}\) and the reference path \(r^\star\):
\[
A_{\mathrm{maze}}=
\frac{\mathrm{LCS}(\hat{r},r^\star)}{\max(1,|r^\star|)}.
\]

\subsection{Tower of Hanoi}

Hanoi samples use \(2\)--\(5\) disks and three rods.  In-domain samples initialize disks on rods \(\{0,1\}\), while OOD samples may initialize disks on \(\{0,1,2\}\) and require at least one disk to already appear on the goal rod.  The goal is always rod \(2\).  For each initial assignment, the generator solves the shortest legal plan with breadth-first search and renders one lifted disk move at a time.

The evaluator decodes disk tracks, infers moves of the form \((\mathrm{disk},\mathrm{source},\mathrm{target})\), and simulates them from the ground-truth initial state.  The success score requires all inferred moves to be legal and the final state to place every disk on the goal rod.  The average progress score measures plan overlap with the reference shortest plan:
\[
A_{\mathrm{hanoi}}=
\frac{2\,\mathrm{LCS}(\hat{m},m^\star)}{|\hat{m}|+|m^\star|}.
\]

\subsection{One-line Drawing}

One-line drawing levels \(2,3,4\) use increasingly larger boards and longer self-avoiding paths.  The generator samples a topology family, searches for an adjacent-cell path without revisits, applies random geometric transforms, checks uniqueness and shape constraints, and treats the resulting path as both the occupied shape and the required solution.  In the evaluation set, level 2 uses a \(9\times9\) board, level 3 uses \(10\times10\), and level 4 uses \(12\times12\).

The evaluator extracts the drawing head, visited cells, white trace cells, and whether the trace remains on the required shape.  The success score requires starting from the highlighted cell, covering every occupied cell, avoiding revisits, avoiding non-adjacent jumps, and never leaving the shape.  A small visual bridge tolerance is allowed when the white trace already connects an apparent short jump.  The average progress score combines coverage, legal-transition ratio, and on-shape ratio:
\[
A_{\mathrm{one}}=
0.5\,C_{\mathrm{cover}}+
0.3\,R_{\mathrm{legal}}+
0.2\,R_{\mathrm{shape}}.
\]

\subsection{Sliding Puzzle}

Sliding puzzle levels \(2\) and \(3\) use \(3\times3\) boards, while level \(4\) uses a \(4\times4\) board.  The goal state is the ordered board with the blank tile in the final position.  For \(3\times3\) puzzles, states are sampled from an exact solvable pool under bucket constraints; for \(4\times4\) puzzles, states are sampled by backward scrambling from the goal.  Each accepted sample is solved optimally and filtered by solution length, family, and diversity constraints.

The evaluator decodes board states from video frames and selects the best legal subsequence starting at the initial state.  The success score requires the subsequence to reach the goal, preserve tile inventory, contain no illegal blank moves, and have an unresolved-frame ratio below \(0.8\).  The average progress score combines normalized Manhattan progress, final-state quality, rule adherence, and observability:
\[
A_{\mathrm{slide}}=
0.45\,P+
0.20\,Q_{\mathrm{final}}+
0.20\,R_{\mathrm{rule}}+
0.15\,R_{\mathrm{obs}}.
\]

\subsection{Sokoban}

Sokoban samples use an \(8\times8\) grid with walls, one player, one box, and one target.  The generator builds layouts from grammar operators, samples valid player and box positions, solves each candidate with shortest-path search, and filters by action length, push count, box-target distance, deadlock-like cells, and diversity.  Higher levels use longer and more constrained layouts.

The evaluator decodes player and box positions and checks whether the recovered state sequence can be explained by legal walking and pushing.  The success score requires the player and box to start correctly, the box to end on the target, no wall crossing or overlap, no illegal pulling or pushing, and an unresolved-frame ratio below \(0.8\).  The average progress score emphasizes box progress toward the target:
\[
A_{\mathrm{sokoban}}=
0.50\,P_{\mathrm{box}}+
0.20\,Q_{\mathrm{final}}+
0.15\,R_{\mathrm{rule}}+
0.15\,R_{\mathrm{obs}}.
\]

\subsection{Water Pouring}

Water pouring samples contain vertical tubes with fixed capacity and colored blocks.  The generator constructs a solved goal state, samples a backward chain of legal reverse pours to obtain an initial state, solves or validates the resulting puzzle, and filters candidates by solution length, buried blocks, color runs, branching factor, and state diversity.  Higher levels increase tube count, color count, capacity, and solution horizon.

The evaluator extracts tube states from stationary frames.  A legal move transfers the complete contiguous top run of one color into an empty tube or onto the same color, limited by remaining destination capacity.  The success score requires the final tube state to be solved, all decoded transitions to be legal, block conservation and capacity constraints to hold, and unresolved stationary frames to remain below \(0.35\).  The average progress score combines rule adherence, progress, final-state quality, and observability:
\[
A_{\mathrm{water}}=
0.45\,R_{\mathrm{rule}}+
0.35\,P+
0.15\,Q_{\mathrm{final}}+
0.05\,R_{\mathrm{obs}}.
\] 

\section{Entropy-Matched versus Fully Denoised Hierarchies}
\label{sec:entropy_matched}

Our final ablation studies how denoising budgets should be allocated across the hierarchy. A naive strategy is to fully denoise every hierarchy level before moving to the next one. In our implementation, this corresponds to assigning 50 denoising steps to all layers. Although intuitive, this strategy ignores the different entropy scales of different hierarchy levels. Coarse layers contain fewer temporal tokens and represent lower-resolution hypotheses, while fine layers contain more tokens and carry more detailed visual entropy. Therefore, forcing all levels to remove the same amount of uncertainty is not well matched to the hierarchical representation.

We instead use an entropy-matched denoising schedule. Let \(N_\ell\) denote the number of tokens at layer \(\ell\), and let \(\tilde{N}_\ell\) denote its effective temporal support. Since a finer layer represents a larger number of temporal degrees of freedom, we assume that the entropy to be removed at layer \(\ell\) should grow with its effective support:
$$
\Delta H_\ell
\propto
\tilde{N}_\ell^{\beta},
$$
where \(\beta\) is a tunable parameter controlling how aggressively denoising budget increases from coarse to fine layers. Larger \(\beta\) allocates more denoising steps to fine layers, while smaller \(\beta\) makes the schedule more uniform.

We convert this entropy allocation into a layer-wise denoising budget:
$$
K_\ell
=
\left\lceil
K_{\max}
\left(
\frac{\tilde{N}_\ell}{\tilde{N}_L}
\right)^{\beta}
\right\rceil,
$$
where \(K_{\max}\) is the maximum denoising budget used by the finest layer. This rule gives fewer denoising steps to coarse layers and more steps to fine layers, matching the intuition that coarse layers should preserve uncertainty while fine layers should resolve it into concrete visual states.

For the 21-frame setting, our hierarchy has actual layer sizes:
$$
N_\ell = [1,2,4,8,16,21].
$$
The last layer is truncated by the video length, but the underlying binary hierarchy has effective temporal supports:
$$
\tilde{N}_\ell = [1,2,4,8,16,32].
$$
Using \(K_{\max}=50\) and \(\beta=0.66\), the entropy-matched rule gives:
$$
K_\ell
=
\left\lceil
50
\left(
\frac{[1,2,4,8,16,32]}{32}
\right)^{0.66}
\right\rceil
=
[5,8,13,20,32,50].
$$
This is the default denoising schedule used by HDR in the main experiments.

This schedule can also be interpreted through the lens of uncertainty preservation. Coarse layers remove only a small fraction of their uncertainty, so they act as flexible high-level hypotheses. Fine layers remove much more entropy, allowing them to instantiate these hypotheses into detailed frame-level latents. In contrast, the All-50 strategy sets
$$
K_1=K_2=\cdots=K_L=50,
$$
which forces every level to remove the same amount of uncertainty regardless of its temporal scale. This prematurely collapses high-level hypotheses and reduces the ability of lower layers to correct or refine them.

Table~\ref{tab:all50} compares the entropy-matched schedule with fully denoised and alternative schedules. The All-50 variant performs worse overall than the entropy-matched schedule: the overall success score drops from 60.29 to 58.38, and the average progress score drops from 89.56 to 88.21. We also include two additional schedules for completeness: a sparse-to-full schedule \([5,5,5,5,5,50]\), which keeps coarse layers minimally denoised before fully denoising the finest layer, and an exponential schedule \([2,4,8,16,32,50]\), which increases the denoising budget more aggressively. Their benchmark results are left blank.

\begin{table*}[t]
\centering
\small
\caption{Entropy-matched versus alternative denoising schedules. The entropy-matched schedule allocates fewer denoising steps to coarse layers and more steps to fine layers according to their entropy scale, achieving the best overall average progress and remaining competitive in overall success. Compared with sparse-to-full and exponential schedules, entropy matching provides a better balance between preserving high-level uncertainty and refining fine-grained video states.}
\setlength{\tabcolsep}{3.5pt}
\renewcommand{\arraystretch}{1.1}
\resizebox{\textwidth}{!}{%
\begin{tabular}{l c c c c c c c c}
\toprule
Denoising Strategy & Schedule & Hanoi & Maze & One-line & Sliding & Sokoban & Water & Overall \\
\midrule
Entropy-matched 
& $[5,8,13,20,32,50]$ 
& 58.75/79.62 & 78.00/97.18 & 70.00/97.84 & 33.33/93.69 & 78.33/99.69 & 43.33/69.34 & 60.29/89.56 \\

All-50 
& $[50,50,50,50,50,50]$ 
& 60.00/79.70 & 76.00/97.18 & 65.00/97.45 & 30.00/94.34 & 80.00/99.60 & 41.67/65.32 & 58.38/88.21 \\

Sparse-to-full 
& $[5,5,5,5,5,50]$ 
& 55.00/74.58 & 42.00/92.86 & 63.33/97.25 & 41.67/93.76 & 76.67/99.33 & 23.33/58.99 & 50.33/86.13 \\

Exponential 
& $[2,4,8,16,32,50]$ 
& 53.75/77.36 & 58.00/95.53 & 73.33/98.03 & 41.67/94.30 & 86.67/98.73 & 46.67/68.26 & 60.02/88.70 \\
\bottomrule
\end{tabular}}
\label{tab:all50}
\end{table*}

\section{Time Complexity Analysis}
\label{sec:complexity}

We analyze the temporal attention complexity of bidirectional diffusion, Streaming AR Diffusion, and \ours{}. 
Let $N$ denote the number of frame-level video tokens and $K$ denote the number of denoising steps. 
We omit constant factors such as hidden dimension, number of heads, model depth, and the fixed number of tokens attended by HDR.

\inlinepara{Bidirectional diffusion}
Bidirectional diffusion updates the full video sequence at every denoising step. 
Each token attends to all $N$ tokens, so one denoising step costs $\mathcal{O}(N^2)$ temporal attention. 
With $K$ denoising steps, the total complexity is:
$$
\mathcal{O}(K N^2).
$$
This dense all-to-all interaction supports global reasoning, but it repeatedly recomputes the full temporal attention map throughout inference.

\inlinepara{Streaming AR Diffusion}
Streaming AR Diffusion generates tokens from left to right. 
When generating token $i$, it attends to all previously generated tokens. 
Although KV caching avoids recomputing previous hidden states, the attention length still grows with time. 
Thus, the total temporal attention cost is:
$$
\mathcal{O}\left(K \sum_{i=1}^{N} i \right)
=
\mathcal{O}(K N^2).
$$
Therefore, Streaming AR Diffusion improves streaming efficiency compared with bidirectional diffusion, but its attention complexity with respect to the number of tokens remains quadratic.

\inlinepara{\ours{}}
HDR generates a hierarchical latent tree instead of a flat sequence. 
For a video with $N$ frame-level tokens, the total number of hierarchy tokens is linear in $N$; for example, a binary tree contains approximately $2N-1$ nodes. 
More importantly, each token attends only to a fixed-size context, including its previous same-layer token, its parent, and neighboring parent tokens. 
Therefore, the temporal attention cost is linear in the number of hierarchy tokens:
$$
\mathcal{O}(K_{\mathrm{avg}} N),
$$
where $K_{\mathrm{avg}}$ denotes the average denoising budget across hierarchy tokens. 
Since the hierarchy contains approximately $2N-1$ tokens for $N$ frame-level tokens, the full attention cost is $\mathcal{O}(K_{\mathrm{avg}}(2N-1))$, which simplifies to $\mathcal{O}(K_{\mathrm{avg}}N)$. This linear-size hierarchy introduces a longer one-time prefill stage, \(16.19\)s for HDR, compared with \(1.48\)s for bidirectional diffusion and \(2.44\)s for CausalForcing; however, the subsequent streaming latency remains \(0.70\)s per latent and much faster than bidirectional diffusion at \(37.92\)s.
In practice, HDR assigns fewer denoising steps to coarse layers, so $K_{\mathrm{avg}}$ is smaller than the full denoising budget used by standard diffusion.

\begin{table}[t]
\centering
\caption{Temporal attention complexity comparison. Bidirectional diffusion uses all-to-all attention, Streaming AR Diffusion attends to the full prefix, while \ours{} attends to a fixed-size context over a linear-size hierarchy.}
\small
\setlength{\tabcolsep}{5pt}
\renewcommand{\arraystretch}{1.15}
\begin{tabular}{l c c c}
\toprule
Method & Attention Pattern & Token Count & Complexity \\
\midrule
Bidirectional & all-to-all & $N$ & $\mathcal{O}(K N^2)$ \\
Streaming AR Diffusion & prefix attention & $N$ & $\mathcal{O}(K N^2)$ \\
\ours{} & fixed sparse attention & $\approx 2N$ & $\mathcal{O}(K_{\mathrm{avg}} N)$ \\
\bottomrule
\end{tabular}
\label{tab:complexity}
\end{table}

This analysis highlights the main computational advantage of HDR. 
Although HDR introduces additional hierarchy tokens, their number grows only linearly with the video length. 
Since each hierarchy token attends to a constant-size context, the overall temporal attention cost remains linear in $N$. 
In contrast, both bidirectional diffusion and Streaming AR Diffusion have quadratic attention complexity with respect to the number of video tokens.

\section{Comparison with SOTA Closed-source Models}

We further compare \ours{} with state-of-the-art closed-source video generation models, including Wan2.6 and Veo. Since these models are only accessible through closed-source inference APIs or released inference interfaces, we cannot modify their architectures, apply our hierarchical training strategy, or fine-tune them on the proposed benchmark data. Therefore, we evaluate them in an off-the-shelf setting using the same prompts and evaluation protocol as our method.

As shown in Table~\ref{tab:comparison_with_closed_source}, off-the-shelf closed-source models achieve limited success under the exact-completion criterion on our reasoning benchmarks. Although they can often generate visually plausible videos, they frequently fail to satisfy the exact multi-step constraints required by these tasks. In contrast, \ours{} is explicitly trained for HDR and achieves substantially higher scores across all tasks. This comparison highlights that strong general-purpose video generation capability does not necessarily translate into reliable multi-step reasoning under irreversible decision-making.

\begin{table*}[t]
\centering
\caption{Comparison with state-of-the-art closed-source video generation models on video reasoning benchmarks. Scores are reported using the success metric and the average progress metric.}
\small
\setlength{\tabcolsep}{3.0pt}
\renewcommand{\arraystretch}{1.1}

\begin{tabular}{l c c c c c c c c c c}
\toprule
Method & Full Attn. & Fine-tuned & Metric & Hanoi & Maze & One-line & Sliding & Sokoban & Water & Overall \\
\midrule\midrule

\multirow[c]{2}{*}{Wan2.6}
& \multirow[c]{2}{*}{\cmark}
& \multirow[c]{2}{*}{\xmark}
& Success & 3.75 & 6.00 & 1.67 & 0.00 & 0.00 & 1.67 & 2.16 \\
&
&
& Avg.\ Progress & 20.15 & 46.93 & 72.13 & 61.38 & 67.56 & 35.50 & 49.06 \\

\midrule

\multirow[c]{2}{*}{Veo}
& \multirow[c]{2}{*}{\cmark}
& \multirow[c]{2}{*}{\xmark}
& Success & 0.00 & 0.00 & 0.00 & 0.00 & 0.00 & 0.00 & 0.00 \\
&
&
& Avg.\ Progress & 0.09 & 24.42 & 66.46 & 0.00 & 0.33 & 0.82 & 14.28 \\

\midrule\midrule

\multirow[c]{2}{*}{\textbf{\ours{}}}
& \multirow[c]{2}{*}{\xmark}
& \multirow[c]{2}{*}{\cmark}
& Success & \textbf{58.75} & \textbf{78.00} & \textbf{70.00} & \textbf{33.33} & \textbf{78.33} & \textbf{43.33} & \textbf{60.29} \\
&
&
& Avg.\ Progress & \textbf{79.62} & \textbf{97.18} & \textbf{97.84} & \textbf{93.69} & \textbf{99.69} & \textbf{69.34} & \textbf{89.56} \\

\bottomrule
\end{tabular}

\label{tab:comparison_with_closed_source}
\end{table*}

\section{Full Table of Ablations}

The main paper visualizes the key trends of the denoising-step, hierarchical-layer, and data-reduction ablations in Figure~\ref{fig:noisy_reduction}, Figure~\ref{fig:layer_reduction}, and Figure~\ref{fig:data_reduction}. For completeness, we provide the corresponding full numerical results in Table~\ref{tab:step_ablation}, Table~\ref{tab:layer_ablation}, and Table~\ref{tab:data_ablation}. Table~\ref{tab:step_ablation} contains the task-level scores behind Figure~\ref{fig:noisy_reduction}, including the full success and average-progress results for the causal baseline, bidirectional baseline, and \ours{} under different denoising budgets. Table~\ref{tab:layer_ablation} provides the task-level breakdown for Figure~\ref{fig:layer_reduction}, organized by the number of active hierarchical layers from 1 to 6, where the 1-layer setting corresponds to the causal baseline. Table~\ref{tab:data_ablation} reports the task-level results behind Figure~\ref{fig:data_reduction}, comparing the bidirectional baseline and \ours{} at different training data scales. All entries are reported as mean $\pm$ standard deviation. For each task-level metric computed over $N$ evaluation samples, we estimate the standard deviation by repeatedly sampling $\lfloor N/2 \rfloor$ examples without replacement, recomputing the mean for 10 trials, and then reporting the population standard deviation across the resulting sample means.

These tables support the same conclusions as the figures while exposing the per-task details. First, the denoising-step ablation shows that \ours{} retains stronger performance under reduced denoising budgets, especially at the one-step setting. Second, the hierarchical-layer ablation shows that shallower variants stay closer to the causal baseline, while adding more layers generally improves performance and makes HDR's reasoning stronger. Third, the data-reduction ablation shows that HDR degrades more gracefully than the bidirectional baseline as the amount of training data decreases. Together, the full tables confirm that HDR's robustness comes from both its hierarchical coarse-to-fine structure and its ability to preserve useful reasoning behavior under limited denoising computation and limited data.

\begin{table*}[t]
\centering
\small
\caption{Denoising step ablation. Values are reported as mean $\pm$ std. \ours{} remains substantially more robust than the causal baseline under reduced inference budgets.}
\setlength{\tabcolsep}{3.5pt}
\renewcommand{\arraystretch}{1.1}

\resizebox{\linewidth}{!}{%
\begin{tabular}{l c c c c c c c c c}
\toprule
Method & Step & Metric & Hanoi & Maze & One-line & Sliding & Sokoban & Water & Overall \\
\midrule\midrule

\multirow[c]{6}{*}{CausalForcing \cite{zhu2026causalforcingautoregressivediffusion}}
& \multirow[c]{2}{*}{50}
& Success & \meanstd{45.00}{6.73} & \meanstd{12.00}{8.08} & \meanstd{48.33}{7.66} & \meanstd{21.67}{5.16} & \meanstd{40.00}{10.69} & \meanstd{38.33}{5.25} & \meanstd{34.22}{7.26} \\
&
& Avg.\ Progress & \meanstd{70.47}{2.76} & \meanstd{55.04}{8.36} & \meanstd{95.15}{1.28} & \meanstd{86.13}{2.62} & \meanstd{82.25}{5.79} & \meanstd{66.94}{3.03} & \meanstd{76.00}{3.97} \\

\cmidrule(lr){2-10}

& \multirow[c]{2}{*}{5}
& Success & \meanstd{38.75}{5.14} & \meanstd{14.00}{11.31} & \meanstd{43.33}{9.26} & \meanstd{20.00}{6.00} & \meanstd{35.00}{10.19} & \meanstd{36.67}{6.83} & \meanstd{31.29}{8.12} \\
&
& Avg.\ Progress & \meanstd{66.31}{2.78} & \meanstd{49.67}{9.17} & \meanstd{94.30}{2.00} & \meanstd{83.73}{2.79} & \meanstd{79.81}{5.85} & \meanstd{66.97}{3.26} & \meanstd{73.47}{4.31} \\

\cmidrule(lr){2-10}

& \multirow[c]{2}{*}{1}
& Success & \meanstd{7.50}{4.36} & \meanstd{10.00}{5.63} & \meanstd{0.00}{0.00} & \meanstd{18.33}{6.75} & \meanstd{28.33}{8.92} & \meanstd{3.33}{1.80} & \meanstd{11.25}{4.57} \\
&
& Avg.\ Progress & \meanstd{39.24}{4.57} & \meanstd{60.29}{9.31} & \meanstd{92.82}{1.00} & \meanstd{79.13}{3.56} & \meanstd{77.45}{5.79} & \meanstd{32.40}{1.70} & \meanstd{63.55}{4.32} \\

\midrule

\multirow[c]{6}{*}{Bidirectional \cite{wan2025wanopenadvancedlargescale}}
& \multirow[c]{2}{*}{50}
& Success & \meanstd{45.00}{6.78} & \meanstd{90.00}{3.12} & \meanstd{81.67}{7.60} & \meanstd{31.67}{6.98} & \meanstd{90.00}{4.21} & \meanstd{21.67}{6.83} & \meanstd{60.00}{5.92} \\
&
& Avg.\ Progress & \meanstd{73.58}{2.95} & \meanstd{99.89}{0.11} & \meanstd{99.26}{0.27} & \meanstd{93.00}{1.56} & \meanstd{100.00}{0.00} & \meanstd{57.08}{3.66} & \meanstd{87.13}{1.42} \\

\cmidrule(lr){2-10}

& \multirow[c]{2}{*}{5}
& Success & \meanstd{41.25}{6.91} & \meanstd{88.00}{2.56} & \meanstd{75.00}{8.79} & \meanstd{35.00}{5.74} & \meanstd{86.67}{4.81} & \meanstd{26.67}{5.85} & \meanstd{58.77}{5.78} \\
&
& Avg.\ Progress & \meanstd{75.01}{3.04} & \meanstd{99.75}{0.15} & \meanstd{98.70}{0.72} & \meanstd{92.45}{1.66} & \meanstd{99.11}{0.89} & \meanstd{55.34}{3.16} & \meanstd{86.73}{1.60} \\

\cmidrule(lr){2-10}

& \multirow[c]{2}{*}{1}
& Success & \meanstd{5.00}{2.78} & \meanstd{0.00}{0.00} & \meanstd{31.67}{9.08} & \meanstd{28.33}{4.35} & \meanstd{41.67}{9.17} & \meanstd{0.00}{0.00} & \meanstd{17.78}{4.23} \\
&
& Avg.\ Progress & \meanstd{43.86}{3.13} & \meanstd{43.87}{4.43} & \meanstd{91.32}{1.42} & \meanstd{94.26}{0.62} & \meanstd{84.79}{5.62} & \meanstd{23.86}{1.53} & \meanstd{63.66}{2.79} \\

\midrule

\multirow[c]{6}{*}{\ours{}}
& \multirow[c]{2}{*}{Full}
& Success & \meanstd{58.75}{4.50} & \meanstd{78.00}{5.25} & \meanstd{70.00}{10.12} & \meanstd{33.33}{7.93} & \meanstd{78.33}{9.67} & \meanstd{43.33}{4.79} & \meanstd{60.29}{7.04} \\
&
& Avg.\ Progress & \meanstd{79.62}{3.04} & \meanstd{97.18}{1.07} & \meanstd{97.84}{0.61} & \meanstd{93.69}{1.47} & \meanstd{99.69}{0.31} & \meanstd{69.34}{2.84} & \meanstd{89.56}{1.56} \\

\cmidrule(lr){2-10}

& \multirow[c]{2}{*}{5}
& Success & \meanstd{58.75}{5.25} & \meanstd{74.00}{6.93} & \meanstd{75.00}{8.96} & \meanstd{36.67}{8.26} & \meanstd{75.00}{10.62} & \meanstd{43.33}{3.54} & \meanstd{60.46}{7.26} \\
&
& Avg.\ Progress & \meanstd{79.38}{3.56} & \meanstd{96.88}{1.20} & \meanstd{97.88}{0.61} & \meanstd{94.02}{1.59} & \meanstd{97.61}{1.19} & \meanstd{67.18}{2.45} & \meanstd{88.83}{1.77} \\

\cmidrule(lr){2-10}

& \multirow[c]{2}{*}{1}
& Success & \meanstd{50.00}{6.05} & \meanstd{0.00}{0.00} & \meanstd{45.00}{12.03} & \meanstd{35.00}{6.15} & \meanstd{76.67}{7.23} & \meanstd{1.67}{1.00} & \meanstd{34.72}{5.41} \\
&
& Avg.\ Progress & \meanstd{74.03}{2.74} & \meanstd{57.10}{9.21} & \meanstd{93.80}{1.11} & \meanstd{95.30}{0.79} & \meanstd{99.81}{0.19} & \meanstd{32.35}{1.89} & \meanstd{75.40}{2.66} \\

\bottomrule
\end{tabular}
}
\label{tab:step_ablation}
\end{table*}

\begin{table}[t]
\centering
\caption{Hierarchical layer ablation. Values are reported as mean $\pm$ std. Results are organized by the number of active hierarchical layers, where the 1-layer setting corresponds to the causal baseline. Adding more layers generally improves reasoning performance, showing that each hierarchical level contributes to HDR's coarse-to-fine reasoning.}
\small
\setlength{\tabcolsep}{3.5pt}
\renewcommand{\arraystretch}{1.1}

\resizebox{\linewidth}{!}{%
\begin{tabular}{l c c c c c c c c}
\toprule
Layers & Metric & Hanoi & Maze & One-line & Sliding & Sokoban & Water & Overall \\
\midrule\midrule

\multirow[c]{2}{*}{1}
& Success & \meanstd{45.00}{6.73} & \meanstd{12.00}{8.08} & \meanstd{48.33}{7.66} & \meanstd{21.67}{5.16} & \meanstd{40.00}{10.69} & \meanstd{38.33}{5.25} & \meanstd{34.22}{7.26} \\
& Avg.\ Progress & \meanstd{70.47}{2.76} & \meanstd{55.04}{8.36} & \meanstd{95.15}{1.28} & \meanstd{86.13}{2.62} & \meanstd{82.25}{5.79} & \meanstd{66.94}{3.03} & \meanstd{76.00}{3.97} \\

\midrule

\multirow[c]{2}{*}{2}
& Success & \meanstd{31.25}{4.19} & \meanstd{44.00}{7.69} & \meanstd{53.33}{11.81} & \meanstd{20.00}{9.03} & \meanstd{51.67}{6.16} & \meanstd{30.00}{6.83} & \meanstd{38.38}{7.62} \\
& Avg.\ Progress & \meanstd{68.05}{2.86} & \meanstd{80.60}{1.57} & \meanstd{97.17}{0.57} & \meanstd{86.83}{1.95} & \meanstd{83.72}{0.71} & \meanstd{59.17}{2.91} & \meanstd{79.26}{1.76} \\

\midrule

\multirow[c]{2}{*}{3}
& Success & \meanstd{38.75}{4.58} & \meanstd{22.00}{8.03} & \meanstd{75.00}{8.24} & \meanstd{30.00}{9.75} & \meanstd{45.00}{7.24} & \meanstd{30.00}{3.67} & \meanstd{40.12}{6.92} \\
& Avg.\ Progress & \meanstd{73.38}{3.27} & \meanstd{84.76}{1.84} & \meanstd{98.55}{0.46} & \meanstd{87.79}{2.37} & \meanstd{93.72}{0.66} & \meanstd{66.35}{2.72} & \meanstd{84.09}{1.89} \\

\midrule

\multirow[c]{2}{*}{4}
& Success & \meanstd{61.25}{7.69} & \meanstd{72.00}{10.64} & \meanstd{80.00}{11.62} & \meanstd{33.33}{9.57} & \meanstd{73.33}{10.71} & \meanstd{35.00}{4.98} & \meanstd{59.15}{9.20} \\
& Avg.\ Progress & \meanstd{81.59}{3.57} & \meanstd{93.33}{3.16} & \meanstd{98.72}{0.60} & \meanstd{91.23}{2.09} & \meanstd{98.71}{3.18} & \meanstd{66.71}{2.65} & \meanstd{88.38}{2.54} \\

\midrule

\multirow[c]{2}{*}{5}
& Success & \meanstd{57.50}{8.22} & \meanstd{76.00}{9.40} & \meanstd{73.33}{8.10} & \meanstd{33.33}{6.57} & \meanstd{80.00}{8.91} & \meanstd{41.67}{5.27} & \meanstd{60.31}{7.74} \\
& Avg.\ Progress & \meanstd{80.70}{5.10} & \meanstd{95.40}{5.90} & \meanstd{98.57}{0.93} & \meanstd{92.65}{1.99} & \meanstd{99.25}{4.23} & \meanstd{69.32}{3.73} & \meanstd{89.31}{3.65} \\

\midrule

\multirow[c]{2}{*}{6}
& Success & \meanstd{58.75}{4.50} & \meanstd{78.00}{5.25} & \meanstd{70.00}{10.08} & \meanstd{33.33}{7.93} & \meanstd{78.33}{9.67} & \meanstd{43.33}{4.79} & \meanstd{60.29}{7.04} \\
& Avg.\ Progress & \meanstd{79.62}{3.04} & \meanstd{97.18}{1.07} & \meanstd{97.84}{0.46} & \meanstd{93.69}{1.48} & \meanstd{99.69}{0.31} & \meanstd{69.34}{2.90} & \meanstd{89.56}{1.54} \\

\bottomrule
\end{tabular}
}
\label{tab:layer_ablation}
\end{table}

\begin{table*}[t]
\centering
\small
\caption{Data reduction ablation. Values are reported as mean $\pm$ std. We train the bidirectional baseline and \ours{} using the full training set, 10\% of the training set, and 2\% of the training set. This experiment evaluates whether the models can learn reasoning rules from limited data.}
\setlength{\tabcolsep}{3.5pt}
\renewcommand{\arraystretch}{1.1}

\resizebox{\linewidth}{!}{%
\begin{tabular}{l c c c c c c c c c}
\toprule
Method & Data Scale & Metric & Hanoi & Maze & One-line & Sliding & Sokoban & Water & Overall \\
\midrule\midrule

\multirow[c]{6}{*}{Bidirectional \cite{wan2025wanopenadvancedlargescale}}
& \multirow[c]{2}{*}{Full}
& Success & \meanstd{45.00}{6.78} & \meanstd{90.00}{3.12} & \meanstd{81.67}{7.60} & \meanstd{31.67}{6.98} & \meanstd{90.00}{4.21} & \meanstd{21.67}{6.83} & \meanstd{60.00}{5.92} \\
&
& Avg.\ Progress & \meanstd{73.58}{2.95} & \meanstd{99.89}{0.11} & \meanstd{99.26}{0.27} & \meanstd{93.00}{1.56} & \meanstd{100.00}{0.00} & \meanstd{57.08}{3.66} & \meanstd{87.13}{1.42} \\

\cmidrule(lr){2-10}

& \multirow[c]{2}{*}{10\%}
& Success & \meanstd{55.00}{5.36} & \meanstd{80.00}{9.94} & \meanstd{80.00}{9.13} & \meanstd{26.67}{7.24} & \meanstd{70.00}{6.67} & \meanstd{20.00}{6.15} & \meanstd{55.28}{7.42} \\
&
& Avg.\ Progress & \meanstd{73.94}{3.00} & \meanstd{95.08}{2.03} & \meanstd{98.70}{0.59} & \meanstd{93.45}{1.08} & \meanstd{99.08}{0.90} & \meanstd{57.20}{3.55} & \meanstd{86.24}{1.86} \\

\cmidrule(lr){2-10}

& \multirow[c]{2}{*}{2\%}
& Success & \meanstd{43.75}{6.74} & \meanstd{20.00}{7.76} & \meanstd{76.67}{9.27} & \meanstd{16.67}{3.27} & \meanstd{23.33}{7.42} & \meanstd{6.67}{3.14} & \meanstd{31.18}{6.27} \\
&
& Avg.\ Progress & \meanstd{72.54}{2.59} & \meanstd{67.76}{6.35} & \meanstd{97.69}{0.66} & \meanstd{92.51}{0.71} & \meanstd{90.11}{3.32} & \meanstd{47.03}{2.39} & \meanstd{77.94}{2.67} \\

\midrule

\multirow[c]{6}{*}{\ours{}}
& \multirow[c]{2}{*}{Full}
& Success & \meanstd{58.75}{4.50} & \meanstd{78.00}{5.25} & \meanstd{70.00}{10.12} & \meanstd{33.33}{7.93} & \meanstd{78.33}{9.67} & \meanstd{43.33}{4.79} & \meanstd{60.29}{7.04} \\
&
& Avg.\ Progress & \meanstd{79.62}{3.04} & \meanstd{97.18}{1.07} & \meanstd{97.84}{0.61} & \meanstd{93.69}{1.47} & \meanstd{99.69}{0.31} & \meanstd{69.34}{2.84} & \meanstd{89.56}{1.56} \\

\cmidrule(lr){2-10}

& \multirow[c]{2}{*}{10\%}
& Success & \meanstd{56.25}{5.78} & \meanstd{64.00}{9.95} & \meanstd{68.33}{8.34} & \meanstd{38.33}{5.86} & \meanstd{80.00}{5.83} & \meanstd{40.00}{5.57} & \meanstd{57.82}{6.89} \\
&
& Avg.\ Progress & \meanstd{79.26}{3.55} & \meanstd{93.63}{2.05} & \meanstd{97.91}{0.55} & \meanstd{94.40}{0.82} & \meanstd{99.92}{0.08} & \meanstd{71.67}{2.17} & \meanstd{89.46}{1.54} \\

\cmidrule(lr){2-10}

& \multirow[c]{2}{*}{2\%}
& Success & \meanstd{60.00}{5.59} & \meanstd{50.00}{13.43} & \meanstd{50.00}{7.98} & \meanstd{26.67}{5.71} & \meanstd{70.00}{7.20} & \meanstd{43.33}{2.49} & \meanstd{50.00}{7.07} \\
&
& Avg.\ Progress & \meanstd{77.71}{3.75} & \meanstd{86.78}{6.54} & \meanstd{96.40}{0.90} & \meanstd{93.76}{0.68} & \meanstd{98.82}{0.47} & \meanstd{68.84}{2.18} & \meanstd{87.05}{2.42} \\

\bottomrule
\end{tabular}
}
\label{tab:data_ablation}
\end{table*}

\section{Visualization of Hierarchical Intermediate Predictions}
\label{sec:hierarchical_visualization}

\begin{figure*}[t]
    \centering
    \includegraphics[width=\textwidth]{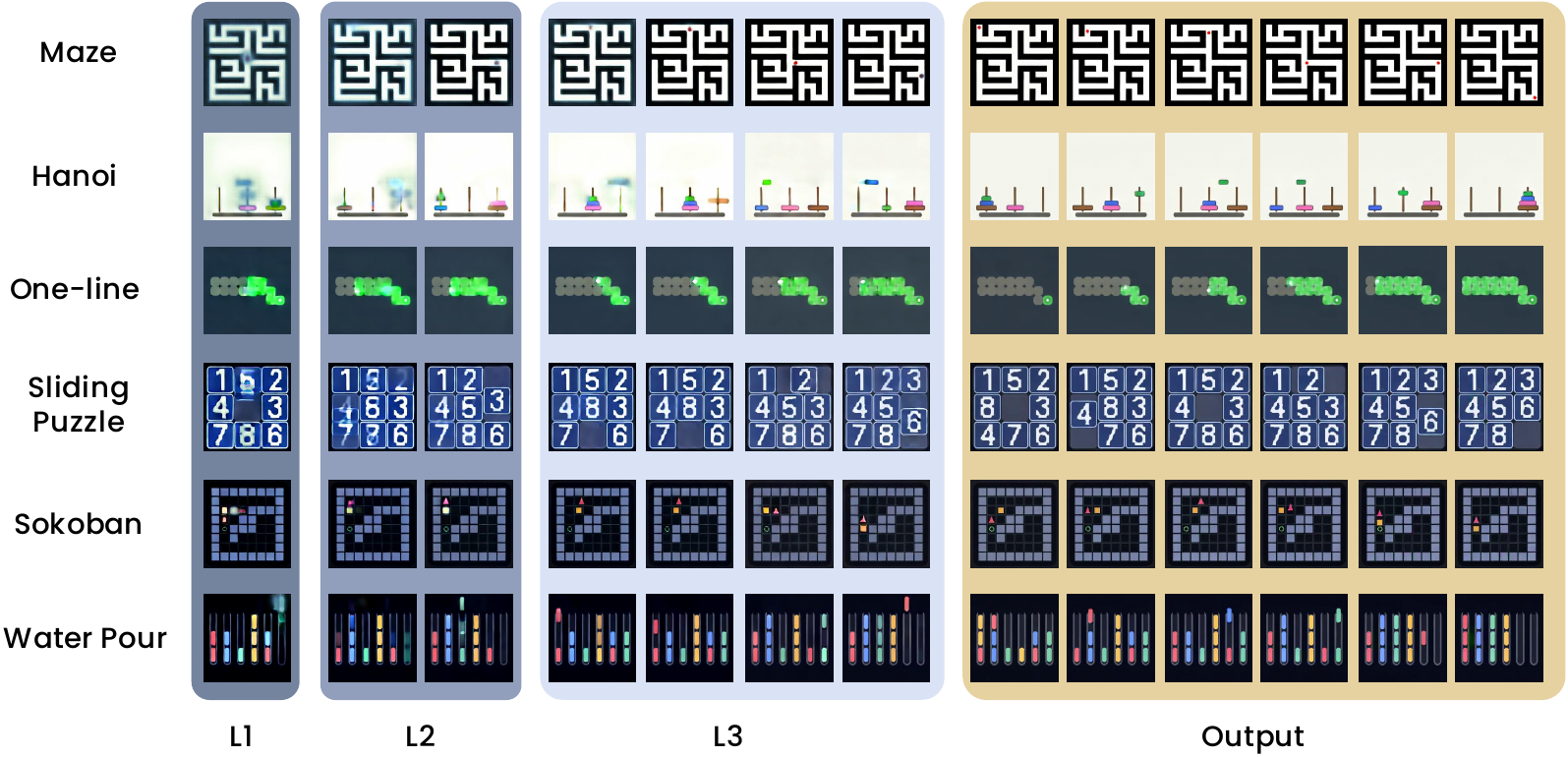}
    \caption{Visualization of HDR's hierarchical intermediate predictions. Columns labeled L1, L2, and L3 show clean \(x_0\) predictions before noise is added at different hierarchy levels, while the rightmost block shows the final generated output. The progression from coarse layers to fine layers illustrates how HDR first forms high-level plans and then refines them into detailed video states.}
    \label{fig:vis}
\end{figure*}

We provide qualitative visualizations of the intermediate predictions produced by \ours{} during hierarchical reasoning. As shown in Figure~\ref{fig:vis}, the columns labeled L1, L2, and L3 correspond to the model's clean \(x_0\) predictions before noise is added at different hierarchy levels. These intermediate outputs reveal the coarse-to-fine inference process of HDR: higher layers first capture coarse task structure and global plans, while lower layers progressively refine them into concrete video states. This supports our claim that HDR enables the model to form high-level hypotheses before committing to final streaming outputs.

\section{Failure Case Studies}
\label{sec:failure_case_studies}

This appendix provides qualitative case studies that complement the quantitative results in the main paper. Section~\ref{sec:qualitative_comparison_streaming_ar} compares HDR with the streaming autoregressive diffusion baseline and illustrates how hierarchical denoising helps avoid early irreversible mistakes. Section~\ref{sec:hdr_residual_failure_modes} then examines representative residual failures of HDR, focusing on state-consistency drift in maze navigation, Sokoban, and water pouring.

\subsection{Qualitative Comparison with Streaming AR Diffusion}
\label{sec:qualitative_comparison_streaming_ar}

We first provide representative qualitative cases where the streaming autoregressive diffusion baseline, represented by CausalForcing, fails while \ours{} succeeds. These examples complement the benchmark results in the main paper by showing a recurring failure mode: the baseline makes an early, locally plausible decision, but its irreversible temporal commitment prevents later frames from repairing the trajectory. Once the rollout deviates from the correct plan, subsequent frames inherit the error and the video ends in a globally inconsistent state.

\begin{figure*}[t]
    \centering
    \includegraphics[width=\textwidth]{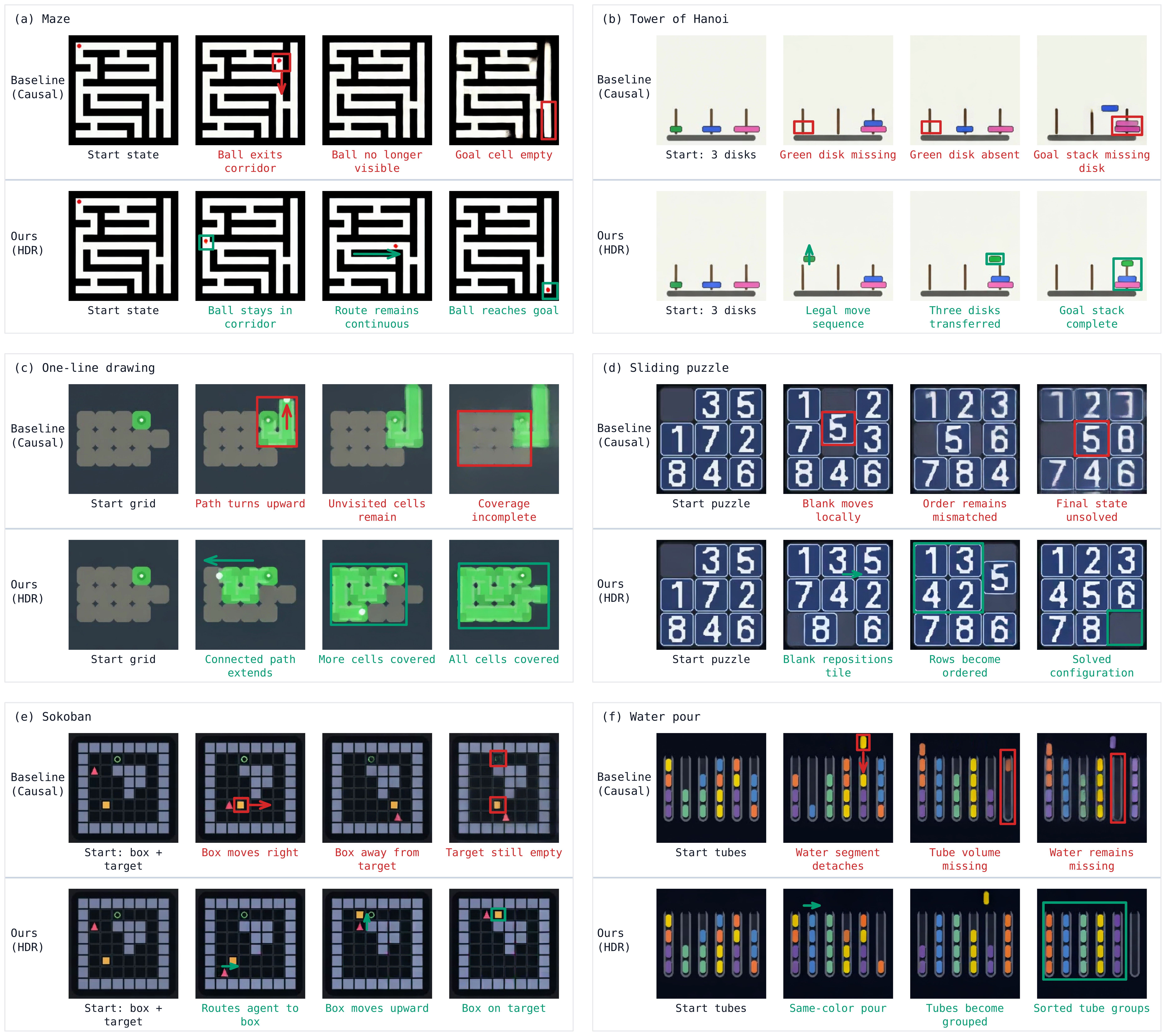}
    \caption{Qualitative comparison across all six reasoning tasks where the streaming autoregressive diffusion baseline fails but \ours{} succeeds. For each task, the top row shows CausalForcing and the bottom row shows \ours{}. Red boxes and arrows mark the baseline's failure points, while green boxes and arrows highlight HDR's successful coarse-to-fine refinement. Across tasks, the baseline typically commits to an early local mistake, such as leaving the valid maze corridor, missing a required disk transfer, breaking path continuity, misplacing the blank tile, pushing the box away from the goal, or violating water-pouring constraints. In contrast, \ours{} maintains globally consistent task structure and reaches the correct final state.}
    \label{fig:case_study_all_tasks}
\end{figure*}

By contrast, \ours{} maintains revisable high-level hypotheses at coarse hierarchy levels and refines them through coarse-to-fine denoising before committing to the final streaming output. This hierarchical inference process enables implicit backtracking in latent space: the model can preserve uncertainty about multi-step structure, revise an incorrect partial plan, and instantiate a rule-consistent solution at finer levels. Figure~\ref{fig:case_study_all_tasks} illustrates this behavior across maze navigation, Tower of Hanoi, one-line drawing, sliding puzzle, Sokoban, and water pouring.

\subsection{Residual Failure Modes of HDR}
\label{sec:hdr_residual_failure_modes}

\begin{figure*}[p]
    \centering
    \includegraphics[width=\textwidth]{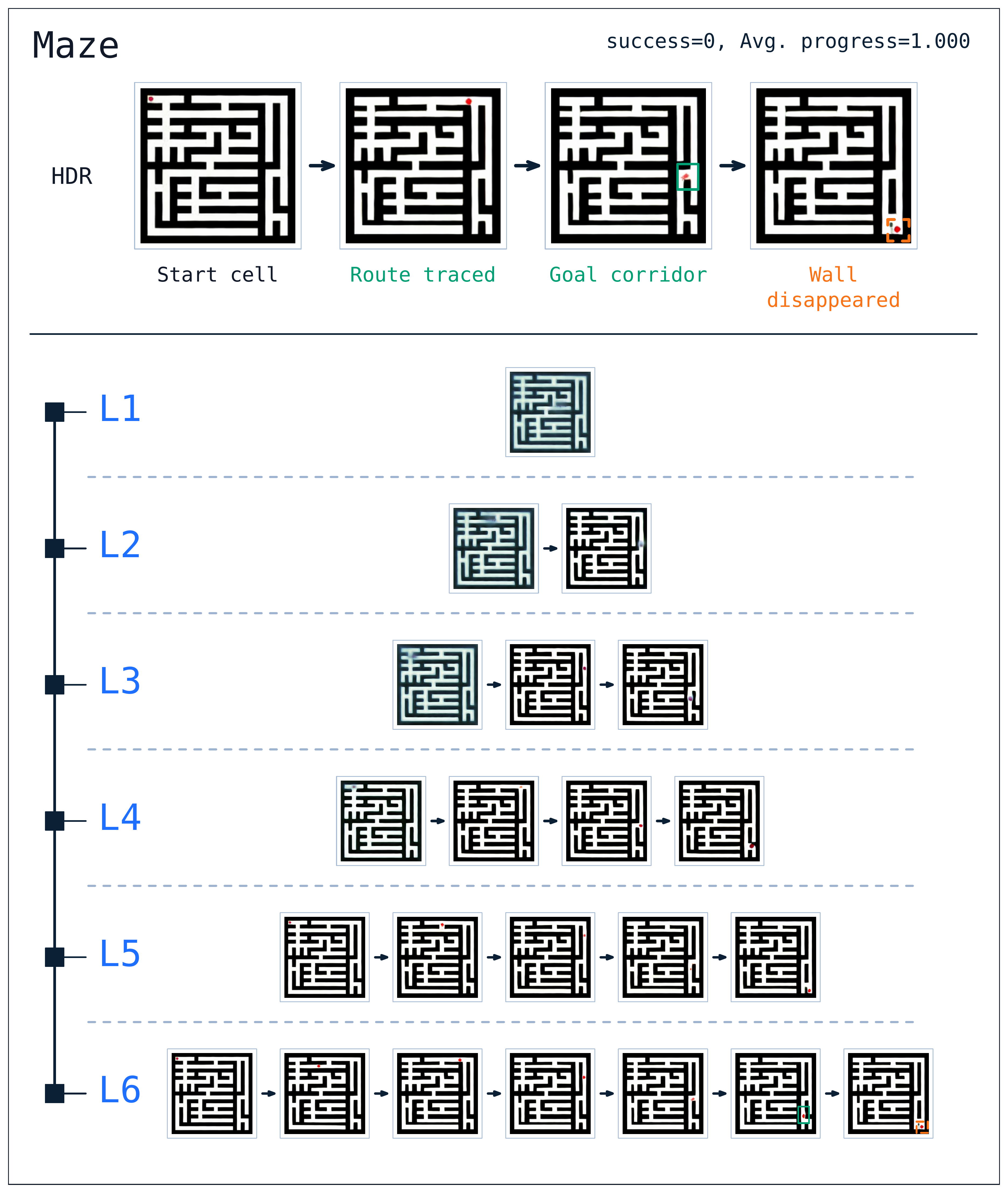}
    \caption{Maze failure case of HDR. The model traces a plausible route and reaches the goal corridor, but a wall disappears near the end of the rollout, producing an invalid maze state despite otherwise correct multi-step planning.}
    \label{fig:maze_failure}
\end{figure*}

\begin{figure*}[p]
    \centering
    \includegraphics[width=\textwidth]{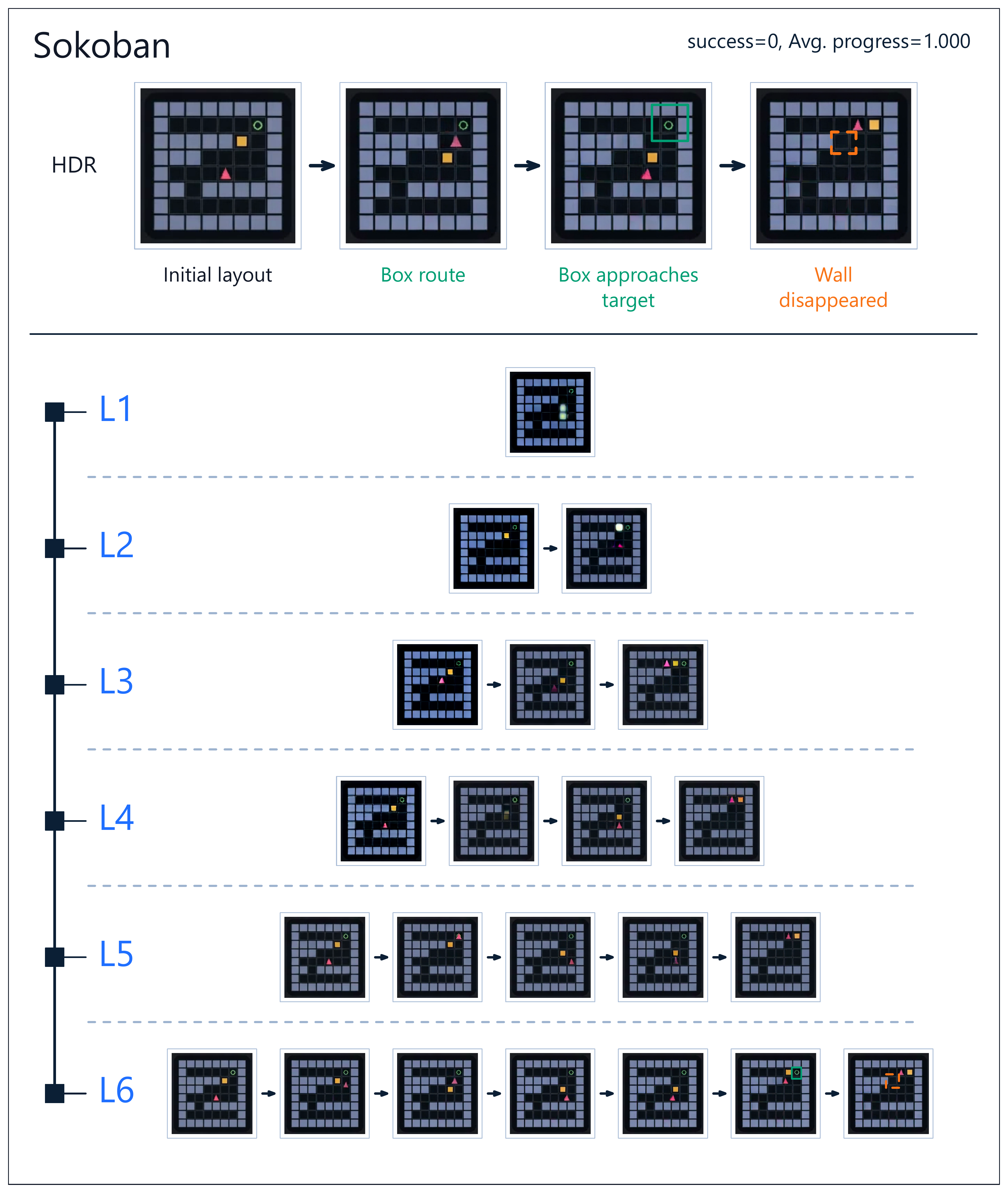}
    \caption{Sokoban failure case of HDR. The box trajectory is largely correct and approaches the target, but a wall disappears near the end of the rollout, making the final scene physically inconsistent.}
    \label{fig:sokoban_failure}
\end{figure*}

\begin{figure*}[p]
    \centering
    \includegraphics[width=\textwidth]{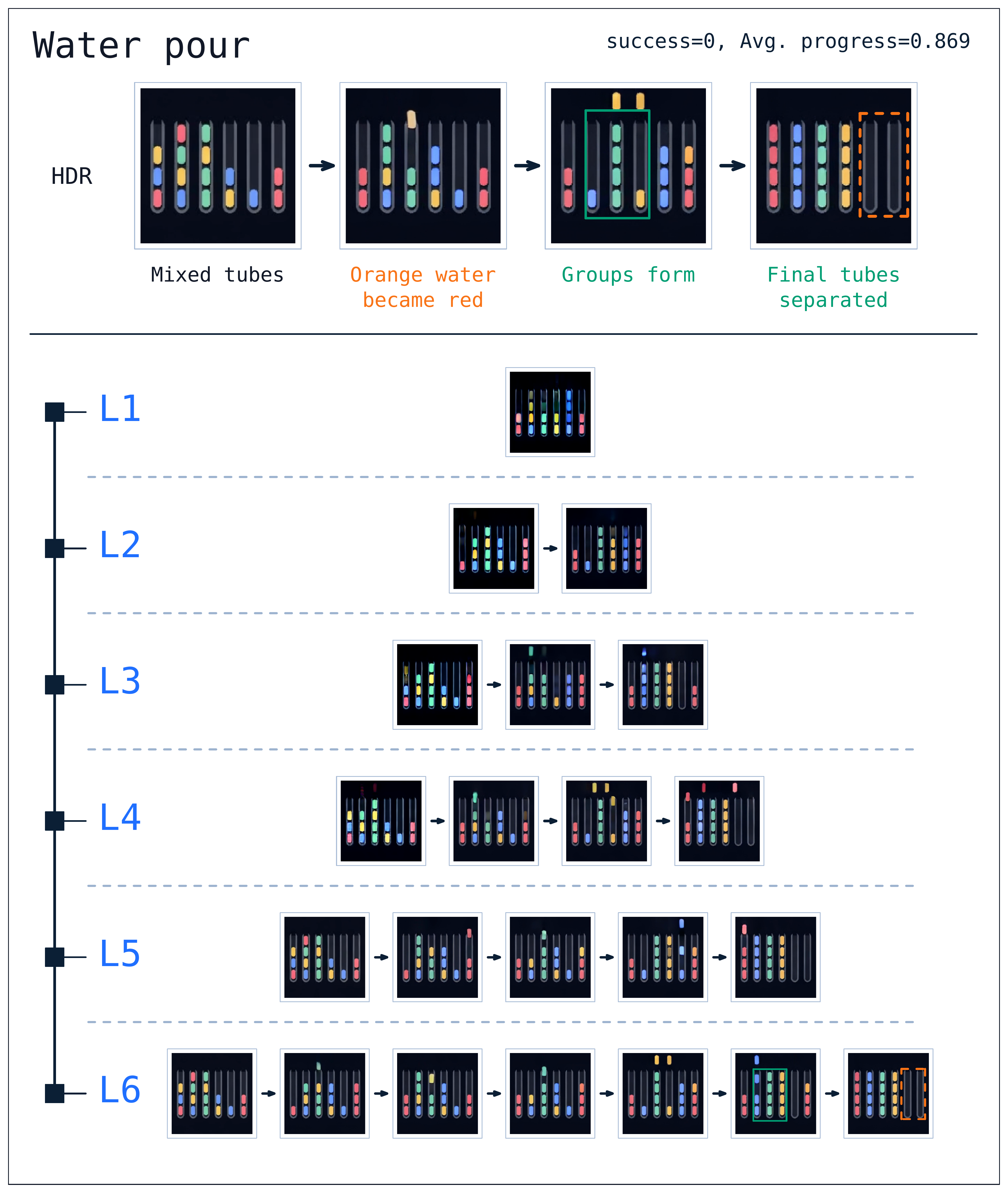}
    \caption{Water-pouring failure case of HDR. The model nearly completes the tube grouping correctly, but orange liquid collapses into red, yielding a visually plausible yet invalid final arrangement.}
    \label{fig:waterpour_failure}
\end{figure*}

Although \ours{} substantially improves multi-step video reasoning, it can still fail when exact state consistency must be preserved until the final frames. Figures~\ref{fig:maze_failure}, \ref{fig:sokoban_failure}, and \ref{fig:waterpour_failure} show three representative residual failure modes. In maze navigation, HDR follows a plausible route but a wall disappears near the end of the rollout, making the final maze state invalid. In Sokoban, the box trajectory largely approaches the target, but a late wall-disappearance artifact breaks physical consistency. In water pouring, the grouping process is mostly correct, yet one liquid color collapses into another, producing a visually plausible but semantically invalid final arrangement.

These failures suggest that HDR's remaining errors are less often complete planning failures and more often late-stage state-consistency drift. The coarse hierarchy levels typically encode the intended multi-step structure, and the finer levels refine this intent into nearly correct video states. The residual errors appear near the end of rollout, where the model must preserve exact geometry, object identity, or terminal constraints. This suggests that future improvements should focus on constraint-aware refinement or lightweight verification during the final generation stage.

\clearpage

\section{Baselines and Implementation Details}
\label{sec:baselines_implementation_details}

\inlinepara{Baselines}
We compare HDR against three families of baselines. The first family consists of full-attention methods, including bidirectional diffusion and VideoMAE~\cite{tong2022videomaemaskedautoencodersdataefficient}. These methods allow global temporal interactions and therefore provide strong reasoning-oriented comparisons, but they require dense sequence processing. For VideoMAE, we enable image-to-video generation by randomly masking \(90\%\)--\(100\%\) of tokens during training and masking all frames except the first frame at inference. The second family consists of non-full-attention methods, including causal diffusion, autoregressive generation, and VideoGPT~\cite{yan2021videogptvideogenerationusing}. These methods are more efficient and better aligned with streaming generation, but they are more vulnerable to irreversible early mistakes. For a controlled comparison, both the bidirectional diffusion baseline and the causal diffusion baseline are built on Wan2.2-5B-TI2V~\cite{wan2025wanopenadvancedlargescale}. The bidirectional baseline follows the original full-attention inference paradigm of the backbone. The causal baseline replaces full temporal attention with a causal attention mask and is trained using the teacher-forcing strategy from Causal Forcing~\cite{zhu2026causalforcingautoregressivediffusion}. This setup isolates the effect of temporal attention and causal rollout while keeping the underlying generative backbone comparable.

\inlinepara{HDR implementation}
Our implementation is also built on Wan2.2-5B-TI2V and augments the backbone with the hierarchical latent tree described in Section~3. For the 125-frame setting used in the main experiments, we employ six hierarchy levels in latent space with sizes \(1,2,4,8,16,32\) and the entropy-matched denoising schedule \([5,8,13,20,32,50]\). Unless otherwise stated, this is the default setting for \ours{}. We provide the derivation and ablation of this entropy-matched schedule in Appendix~\ref{sec:entropy_matched}. For training, we use a mixed reasoning dataset containing 18,000 video clips in total, consisting of 3,000 synthetic videos for each of the six tasks. All videos are resized to \(224\times224\). We train from raw videos rather than precomputed latents: videos are decoded online and encoded by the Wan2.2 TI2V VAE during training. We condition on the first frame and allow this condition to be visible to all hierarchy levels. The model is initialized from a pretrained Wan2.2-5B-TI2V checkpoint and optimized with the flow-matching objective for 100,000 training steps. We use AdamW with learning rate \(2\times10^{-6}\), betas \((0,0.999)\), weight decay \(0.01\), bfloat16 mixed-precision training, gradient checkpointing, and hybrid full-sharding FSDP. The per-device batch size is 1 and the total batch size is 8. During inference, we use classifier-free guidance with scale 3.0. Each main training run uses 8 GPUs and takes approximately 48 hours. All baselines are trained and evaluated under the same configuration for fair comparison.

\end{document}